\documentclass[11pt]{article}
\usepackage[margin=1in]{geometry}
\usepackage{amsmath, amssymb, amsthm}
\usepackage{enumitem}
\usepackage{xcolor}
\usepackage{hyperref}
\usepackage[round]{natbib}
\usepackage{algorithm}
\usepackage{algorithmic}
\usepackage{booktabs}
\usepackage{setspace}
\usepackage{graphicx}
\usepackage{authblk}
\usepackage{subcaption}
\newcommand{\E}{\mathbb{E}}

\newcommand{\X}{\mathbf{X}}

\theoremstyle{definition}

\newtheorem{Remark}{Remark}
\newtheorem{Assumption}{Assumption}

\newtheorem*{Theorem*}{Theorem}
\newtheorem*{Assumption*}{Assumption}

\title{Shapley Value Estimation for Multi-Site Data with Blockwise-Missing Features}

\author[1]{Siqi Li\thanks{siqili@u.duke.nus.edu}}
\author[2]{Wangxuan Fan}
\author[3]{Yiming Li}
\author[4]{Doudou Zhou}
\author[5,6]{Molei Liu}

\affil[1]{Centre for Biomedical Data Science, Duke-NUS Medical School}
\affil[2]{School of Data Science, Chinese University of Hong Kong, Shenzhen}
\affil[3]{Department of Biostatistics, Columbia University}
\affil[4]{Department of Statistics and Data Science, National University of Singapore}
\affil[5]{Department of Biostatistics, Peking University}
\affil[6]{Beijing International Center for Mathematical Research, Peking University}
\date{}

\begin{document}
\maketitle
\vspace{-3em}

\begin{abstract}
Shapley value (SV)-based methods are the prevailing framework for feature attribution in machine learning, yet existing population-level Shapley estimators generally assume that observations used to evaluate the coalitional game are fully observed under a common feature space. 
This assumption is routinely violated in multi-site studies across biomedicine, social science, and environmental monitoring, where institutions record different features under different protocols, producing systematic blockwise missingness across sources.
We first show that the standard remedy of imputing missing features before computing Shapley values introduces systematic, coalition-dependent bias into the resulting attributions. 
We then propose \textbf{FUSHAP} (\textbf{Fu}sion \textbf{Sh}apley \textbf{A}ttribution from \textbf{P}artially-observed data), a method that leverages partially-observed auxiliary sites to reduce the variance of a preliminary single-site Shapley estimate without imputation. 
A permutation-based screening step detects and excludes sites whose data distributions are incompatible with the target population. 
In synthetic experiments, FUSHAP achieves $3$--$8\times$ lower MSE than the single-site estimator and $2$--$3\times$ lower MSE than imputation baselines without incurring imputation-induced bias, and the screening procedure identifies misaligned sites with $82\%$ power at moderate misalignment and $100\%$ for strong misalignment.
On multi-site air quality and multi-center clinical data, FUSHAP reduces MSE by approximately $3$--$7\times$ relative to the single-site estimator; in the clinical application, standard imputation can increase MSE above the single-site baseline.
\end{abstract}

\vspace{0.1em}

\section{Introduction}
\label{sec.intro}
Model interpretability has become an important component of trustworthy machine learning, particularly in high-stakes domains such as medicine, credit scoring, and criminal justice where practitioners must understand why a model produces a given prediction before acting on it. 
Among approaches to post-hoc explanation, Shapley value (SV)-based feature attribution methods, exemplified by SHAP~\citep{lundberg2017shap} and SAGE~\citep{covert2020sage}, have emerged as the principled standard, owing to their axiomatic foundation in cooperative game theory and model-agnostic applicability~\citep{molnar2020interpretable, mosca2022shap,li2024shapley,salih2025perspective}. 
Given $p$ input features and a predictive model $f$, SV methods define a cooperative game $v(S)$ that measures the predictive performance of $f$ when only features in $S \subseteq \{1, \ldots, p\}$ are available, with absent features marginalized over a reference distribution. 
The SV $\phi_j$ of feature $j$ is then the weighted average of $j$'s marginal contribution $v(S \cup \{j\}) - v(S)$ across all $2^{p-1}$ coalitions not containing $j$; estimation of $\boldsymbol\phi = (\phi_1, \ldots, \phi_p)$ requires evaluating $v(S)$ for many coalitions against a reference sample.

A fundamental assumption underlying most existing population-level SV estimators, including KernelSHAP~\citep{lundberg2017shap, covert2021kernelshap}, SAGE~\citep{covert2020sage}, FastSHAP~\citep{jethani2022fastshap}, and SIM-Shapley~\citep{fan2025simshapley}, is that this reference sample is drawn from a single, fully-observed data source. 
In practice, any scientific study that integrates data across multiple sources, whether clinical registries, sensor networks, or multi-cohort social surveys, risks violating this assumption, as different sources typically measure different variables and produce \emph{blockwise missingness} in which entire groups of features are systematically absent at certain sites. 
Clinical research provides a particularly prominent example: large-scale healthcare consortia aggregate records from dozens of institutions, each administering different diagnostic protocols, so that certain imaging, laboratory, or cognitive assessments are entirely unavailable at certain centers~\citep{li2025defuse,li2026distributionallyrobusttransferlearning}. 
To our knowledge, existing SV methods do not explicitly address data partitioned across sites with disjoint blocks of unobserved features and potential distributional shift.

Two natural strategies exist for computing SV from blockwise-missing multi-site data, but neither is fundamentally adequate. 
The most straightforward approach is to compute feature-level SV independently at each site and average the results, which ignores distributional differences across sites and provides no mechanism for detecting sources whose data are incompatible with the target population. 
The other natural remedy is to impute the missing features and proceed with standard SV estimation, but this can be more problematic: imputation alters the covariance structure on which Shapley attributions depend, introducing method-dependent bias into feature importance rankings that does not necessarily diminish with improved predictive accuracy~\citep{vo2025explainabilitymachinelearningmodels}. 

\subsection{Related Work}
\paragraph{Shapley values under incomplete data.}
Computing SV requires specifying how absent features are handled within each coalition~$S$; the choice among conditional, marginal, and baseline removal strategies materially affects the resulting attributions~\citep{chen2023algorithms,covert2021explaining}, and which convention is preferable remains open and context-dependent.
The majority of recent methodological work has focused on improving the computational efficiency of SV estimation under a fixed removal
convention~\citep{covert2021kernelshap, fan2025simshapley, mitchell2022sampling}, leaving the statistical challenge of heterogeneous, partially-observed evaluation data largely unaddressed.

A compounding difficulty arises when the evaluation data themselves contain missing entries. Under missingness, standard imputation yields a surrogate distribution $\widetilde{P}_X \neq P_X$ that does not correct for the shift between observed and full data~\citep{shannon2026distributionshiftmissingdata,
naf2026goodimputationmarmissingness}. Since all common SV formulations define the coalitional game through expectations with respect to~$P_X$, this distributional error propagates directly into the value function~$\mathcal{V}(S)$ and hence into every attribution.
Empirically,~\citet{vo2025explainabilitymachinelearningmodels} confirm that different imputation strategies produce systematically divergent Shapley attributions. Yet no existing work provides a correction for this bias under structured blockwise missingness.

\paragraph{Data fusion under blockwise missingness.}
Several recent works address estimation from multi-source data in which different sources observe different variable subsets.
\citet{xue2021mbi} integrate multiple conditional-mean imputations, each derived from a distinct overlap of observed covariates across block-wise missing-pattern groups, within a penalized generalized method of moments (GMM).
\citet{jin2023modularregressionimprovinglinear} propose a modular regression framework that leverages auxiliary variables satisfying a conditional independence structure to improve estimation efficiency and prediction accuracy.
\citet{li2025defuse} develop a data-adaptive control-variate framework that handles both blockwise missingness and distributional shift for generalized linear model coefficients.
\citet{xu2025blockwisemissingnessmeetsai} and~\citet{huang2025efficientsemiparametricinferencedistributed} extend similar ideas to broader parameter classes under block-missing designs.

In all such cases, the target estimand is defined by a single estimating equation or a small system of moment conditions. 
The Shapley attribution vector $\boldsymbol\phi \in \mathbb{R}^p$, while also finite-dimensional, is defined through $2^p$ coalition-level value functions, each involving a separate conditional expectation, which is a structure absent from prior data-fusion targets. How to extend variance-reduction techniques from scalar estimands to this combinatorial setting remains an open problem.

\paragraph{Shapley values in multi-site and federated settings.}
Several works employ Shapley values in multi-site contexts, but target fundamentally different estimands from ours. One active line assigns a single Shapley value to each \emph{site}, i.e., quantifying how much each data source contributes to the overall model, rather than to each \emph{feature} within the model~\citep{wang2020principledapproachdatavaluation,
Zheng_2023, liu2022gtgshapley}. 
This is client-level data valuation: the players in the cooperative game are institutions, not input variables, and blockwise missingness plays no role. \citet{wu2021explainingmedicalaiperformance} use SV to explain performance disparities across clinical sites, but treat site-level confounders (demographics, equipment type) as the players rather than model features. In all of these formulations, each site has access to the same feature space; the heterogeneous feature coverage that defines blockwise missingness is absent.

\subsection{Contributions}
We propose \textbf{FUSHAP} (\textbf{Fu}sion \textbf{Sh}apley \textbf{A}ttribution from \textbf{P}artially-observed data), a framework for estimating Shapley feature attributions from multi-site data with blockwise-missing covariates, without resorting to imputation. 
Our contributions are as follows.

\begin{enumerate}
\item \textbf{Imputation bias in blockwise-missing settings.} 
Extending the empirical findings of~\citet{vo2025explainabilitymachinelearningmodels},
we confirm that imputing missing features before computing Shapley values introduces systematic, coalition-dependent bias that persists across the standard imputation methods considered, with MSE up to $3.1\times$ that of imputation-free alternatives.
\item \textbf{Variance-reduced estimation without imputation.} 
We derive the influence function of the constrained WLS Shapley estimator and use it to construct control variate corrections from blockwise-missing auxiliary sites, reducing variance without imputing unobserved features.
\item \textbf{Adaptive source screening and calibration.} 
We develop a permutation-based screening procedure that detects incompatible sites, and a total-variance calibration that optimally weights each site's contribution.
\item \textbf{Empirical validation.} 
Across simulations and multi-site real data, FUSHAP reduces MSE by $3$--$8\times$ relative to the single-site estimator and $2$--$3\times$ relative to imputation baselines in simulations, with approximately $3$--$7\times$ improvements over the single-site estimator on real data.
\end{enumerate}

\section{Problem Formulation} \label{sec:formulation}

\subsection{Data structure}\label{sec.data}

Let $Y$ denote the outcome of interest and $\mathbf{X} = (X_1, \ldots, X_p)^\top$ a $p$-dimensional feature vector. We consider a multi-site setting with three types of data source:
\begin{itemize}
\item \emph{Labeled complete} ($\mathcal{LC}$), of size~$n$: both $(Y, \mathbf{X})$ are jointly observed.
\item \emph{Labeled missing} ($\mathcal{LM}_r$, $r = 1, \ldots, R$), of size~$n_r$: the outcome $Y$ and a subset $\mathbf{X}_{\Gamma_r}$ of covariates are observed, where $\Gamma_r \subsetneq \{1, \ldots, p\}$. The remaining covariates $\mathbf{X}_{\Gamma_r^c}$ are entirely unobserved.
\item \emph{Unlabeled complete} ($\mathcal{UC}$), of size~$N \gg n$: all covariates $\mathbf{X}$ are observed but $Y$ is unavailable.
\end{itemize}

The $\mathcal{UC}$ sample defines the target population on which inference is desired. 
We assume centralized access to row-level data from all sources. Throughout, we denote by $\mathcal{D} = \{1, \ldots, p\}$ the full index set and $\rho_r = n_r / n$ the sample-size ratio of the $r$-th labeled-missing source to the complete source.
Sources may differ in their marginal covariate distributions; identification relies on conditional alignment of the outcome and remaining features given an observed alignment set, formalized below.

\begin{Assumption}[Missing at random with sufficient alignment]
\label{asm:mar}
For each site $\mathcal{LM}_r$, $r = 1, \ldots, R$, there
exists a sufficient alignment set
$\Omega_r \subseteq \Gamma_r$ such that
\begin{equation}
  p_{\mathcal{UC}}\!\left(Y, \mathbf{X}_{\Omega_r^c}
    \mid \mathbf{X}_{\Omega_r}\right)
  \;=\;
  p_{\mathcal{LM}_r}\!\left(Y, \mathbf{X}_{\Omega_r^c}
    \mid \mathbf{X}_{\Omega_r}\right).
  \label{eq:mar}
\end{equation}
\end{Assumption}

Assumption~\ref{asm:mar} requires that, conditional on the alignment variables $\mathbf{X}_{\Omega_r}$, the joint distribution of the outcome and remaining features is the same at site~$r$ and in the target population. This generalizes the missing-completely-at-random (MCAR) condition commonly adopted in the blockwise-missing literature~\citep{xue2021mbi,jin2023modularregressionimprovinglinear}: when $\Omega_r = \varnothing$, \eqref{eq:mar} reduces to MCAR; when $\Omega_r = \Gamma_r$, arbitrary marginal shift in $\mathbf{X}$ is permitted provided the conditional distributions agree~\citep{li2025defuse}.

\subsection{Shapley feature attribution}
\label{sec.estimand}
Let $f : \mathbb{R}^p \to \mathbb{R}$ be a fixed, pre-trained predictive model and $\ell : \mathbb{R} \times \mathbb{R} \to \mathbb{R}_{\geq 0}$ a loss function. For each coalition $S \subseteq \mathcal{D}$, define the restricted prediction
\begin{equation}
  \bar{f}_S(\mathbf{x}_S)
  \;=\;
  \mathbb{E}_{p_{\mathcal{UC}}}
  \!\big[f(\mathbf{x}_S, \mathbf{X}_{S^c})\big],
  \label{eq:restricted-pred}
\end{equation}
which marginalizes the absent features $\mathbf{X}_{S^c}$ over their marginal distribution under $p_{\mathcal{UC}}$, independently of~$\mathbf{X}_S$. 
This is the marginal feature removal convention~\citep{lundberg2017shap, fan2025simshapley}; see Remark~\ref{rem:removal} for the conditional alternative.

The SAGE cooperative game~\citep{covert2020sage} assigns to each coalition $S \subseteq \mathcal{D}$ the value
\begin{equation}
  \mathcal{V}(S)
  \;=\;
  -\,\mathbb{E}_{p_{\mathcal{UC}}}
  \!\big[\,
    \ell\!\big(
      \bar{f}_S(\mathbf{X}_S),\; Y
    \big)
  \big],
  \label{eq:value-fn}
\end{equation}
the negated expected loss under coalition~$S$, with larger values indicating better predictive performance. Encoding coalitions as binary vectors $\mathbf{z} \in \{0,1\}^p$ via $S(\mathbf{z}) = \{j : z_j = 1\}$, we write $\mathcal{V}(\mathbf{z})$ and $\mathcal{V}(S)$ interchangeably. 
The population value function admits a per-observation decomposition $\mathcal{V}(\mathbf{z}) = \mathbb{E}_{p_{\mathcal{UC}}}
[\eta(\mathbf{z}, \mathbf{X}, Y)]$, where
\begin{equation}
  \eta(\mathbf{z}, \mathbf{x}, y)
  \;=\;
  -\,\ell\!\big(
    \bar{f}_{S(\mathbf{z})}(\mathbf{x}_{S(\mathbf{z})}),
    \;y
  \big)
  \label{eq:eta}
\end{equation}
records the negated loss for a single observation $(\mathbf{x}, y)$ under coalition~$\mathbf{z}$.

The SV of feature $j$ is the weighted average of its marginal contribution $\mathcal{V}(S \cup \{j\}) - \mathcal{V}(S)$ over all coalitions $S \not\ni j$, uniquely characterized by the efficiency, symmetry, linearity, and null-player axioms~\citep{covert2020sage}. 
Equivalently, $\bar{\boldsymbol\phi} = (\phi_1, \ldots, \phi_p)^\top$ is the solution to the constrained weighted least squares (WLS) problem~\citep{lundberg2017shap,
covert2021kernelshap}
\begin{equation}
  \bar{\boldsymbol\phi}
  \;=\;
  \arg\min_{\boldsymbol\beta \in \mathbb{R}^p}\;
  \mathbb{E}_{\mu_{\mathrm{Sh}}}
  \Big[
    \big(
      \mathcal{V}(\mathbf{0})
      + \mathbf{z}^\top\boldsymbol\beta
      - \mathcal{V}(\mathbf{z})
    \big)^2
  \Big]
  \quad\text{s.t.}\quad
  \mathbf{1}^\top\boldsymbol\beta
  = \mathcal{V}(\mathbf{1}) - \mathcal{V}(\mathbf{0}),
  \label{eq:wls}
\end{equation}
where $\mu_{\mathrm{Sh}}$ is the Shapley kernel, the distribution over coalitions of intermediate size ($0 < |\mathbf{z}| < p$) with probability mass $\mu_{\mathrm{Sh}}(\mathbf{z}) \propto [\binom{p}{|\mathbf{z}|}\,|\mathbf{z}|\, (p - |\mathbf{z}|)]^{-1}$, and the efficiency constraint ensures that the attributions sum to the difference between full-model and null-model performance. 
The KKT conditions yield~\citep{covert2021kernelshap, fan2025simshapley}
\begin{equation}
  \bar{\boldsymbol\phi}
  \;=\;
  \Sigma^{-1}
  \bigg[
    \mathbf{b}
    + \mathbf{1}\,
    \frac{c - \mathbf{1}^\top\Sigma^{-1}\mathbf{b}}
         {\mathbf{1}^\top\Sigma^{-1}\mathbf{1}}
  \bigg],
  \label{eq:kkt}
\end{equation}
with
\begin{equation}
  \Sigma
  = \mathbb{E}_{\mu_{\mathrm{Sh}}}
    [\mathbf{z}\mathbf{z}^\top],
  \qquad
  \mathbf{b}
  = \mathbb{E}_{\mu_{\mathrm{Sh}}}
    \big[\mathbf{z}\big(\mathcal{V}(\mathbf{z})
    - \mathcal{V}(\mathbf{0})\big)\big],
  \qquad
  c = \mathcal{V}(\mathbf{1})
    - \mathcal{V}(\mathbf{0}).
  \label{eq:Sigma-b-c}
\end{equation}
In practice, the expectation over the Shapley kernel is approximated using $m$ sampled coalitions $\mathbf{z}_1,\ldots,\mathbf{z}_m \overset{\mathrm{i.i.d.}}{\sim} \mu_{\mathrm{Sh}}$, where $m$ denotes the coalition-sampling budget.
Let $A = \frac{1}{m}\sum_{j=1}^m \mathbf{z}_j\mathbf{z}_j^\top$, 
$\bar{\mathbf{b}} = \frac{1}{m}\sum_{j=1}^m
  \mathbf{z}_j(\widehat{\mathcal{V}}(\mathbf{z}_j)
  - \widehat{\mathcal{V}}(\mathbf{0}))$, 
  and $\widehat{c} = \widehat{\mathcal{V}}(\mathbf{1}) - \widehat{\mathcal{V}}(\mathbf{0})$.
Note that $A \xrightarrow{p} \Sigma$ and $\bar{\mathbf{b}} \xrightarrow{p} \mathbf{b}$ as $m \to \infty$.

Since $\Sigma$ depends only on the Shapley kernel (not on the data), the data-dependent part of $\bar{\boldsymbol\phi}$ enters entirely through $\mathbf{b}$ and $c$. Both are expectations over the $\mathcal{UC}$ population:
\begin{align}
\mathbf{b}
&= \E_{\mathbf{z}}\big[
     \mathbf{z}\,
     \E_{(\X,Y) \sim p_{\mathcal{UC}}}[\eta(\mathbf{z}, \X, Y)]
   \big]
 - \mathcal{V}(\mathbf{0})\,\E_{\mathbf{z}}[\mathbf{z}],
\label{eq:bbar-expand}
\\
c &= \E_{(\X,Y) \sim p_{\mathcal{UC}}}[\eta(\mathbf{1},\X,Y)
     - \eta(\mathbf{0},\X,Y)].
\label{eq:c-expand}
\end{align}

The estimand $\bar{\boldsymbol\phi}$ is defined via expectations under $p_{\mathcal{UC}}$, but evaluating $\eta(\mathbf{z}, \mathbf{x}, y)$ requires both the outcome~$Y$ and all features~$\mathbf{X}$. Only $\mathcal{LC}$ possesses both, yet its sample size~$n$ is typically small, yielding a high-variance estimate, and its covariate distribution may differ from the target population~$p_{\mathcal{UC}}$, introducing bias.
The $\mathcal{LM}_r$ sources provide additional labeled observations but lack the features in~$\Gamma_r^c$; the $\mathcal{UC}$ source provides the complete feature vector but no outcome. The central question addressed in this paper is whether these partially-observed data sources can reduce the variance of the $\mathcal{LC}$-only estimator without introducing bias.

\begin{Remark}[Feature removal convention]
\label{rem:removal}
Equation~\eqref{eq:restricted-pred} adopts the marginal removal convention, in which $\mathbf{X}_{S^c}$ is drawn independently of~$\mathbf{X}_S$. The conditional alternative $\bar{f}_S(\mathbf{x}_S) = \mathbb{E}_{p_{\mathcal{UC}}}[f(\mathbf{X}) \mid \mathbf{X}_S = \mathbf{x}_S]$ preserves feature dependencies but requires estimating high-dimensional conditional distributions. The estimation framework in Section~\ref{sec.method} is agnostic to this choice: it requires only that $\mathcal{V}(\mathbf{z}) = \mathbb{E}_{p_{\mathcal{UC}}} [\eta(\mathbf{z}, \mathbf{X}, Y)]$ for some per-observation function~$\eta$, a property satisfied under either convention.
\end{Remark}

\section{Method}\label{sec.method}
The goal is to estimate the population Shapley vector $\bar{\boldsymbol\phi}(p_{\mathcal{UC}}, f)$, defined with respect to the $\mathcal{UC}$ covariate distribution and the fixed, pre-trained model~$f$, using data from the three source types described in Section~\ref{sec.data}.

\subsection{Preliminary estimator from the complete-data site}
Given $m$ coalitions $\mathbf{z}_1, \ldots, \mathbf{z}_m$ drawn from the Shapley kernel $\mu_{\mathrm{Sh}}$, the value function $\mathcal{V}(\mathbf{z})$ is estimated from $\mathcal{LC}$ by the importance-weighted sample average
\begin{equation}
  \widehat{\mathcal{V}}(\mathbf{z})
  \;=\;
  \frac{1}{n}\sum_{i=1}^{n}
  \widehat{w}(\mathbf{x}_i)\;
  \eta(\mathbf{z},\, \mathbf{x}_i,\, y_i),
  \label{eq:v-hat}
\end{equation}
where the summation runs over $\mathcal{LC}$ observations.
The density ratio $\widehat{w}(\mathbf{x}) = \widehat{p}_{\mathcal{UC}}(\mathbf{x}) \,/\, \widehat{p}_{\mathcal{LC}}(\mathbf{x})$ reweights
$\mathcal{LC}$ to the target distribution $p_{\mathcal{UC}}$ and is estimated separately by training a binary classifier on $\mathcal{LC} \cup \mathcal{UC}$ with source indicators; $\widehat{w} \equiv 1$ when no covariate shift is present.
The preliminary Shapley estimator $\widetilde{\boldsymbol\phi}$ is then the closed-form solution~\eqref{eq:kkt} with $(\Sigma, \mathbf{b}, c)$ replaced by their sample analogues
\begin{equation}
  \widehat{\Sigma}
  = \tfrac{1}{m}\textstyle\sum_{j=1}^{m}
    \mathbf{z}_j \mathbf{z}_j^\top,
  \quad
  \widehat{\mathbf{b}}
  = \tfrac{1}{m}\textstyle\sum_{j=1}^{m}
    \mathbf{z}_j\big(
    \widehat{\mathcal{V}}(\mathbf{z}_j)
    - \widehat{\mathcal{V}}(\mathbf{0})\big),
  \quad
  \widehat{c}
  = \widehat{\mathcal{V}}(\mathbf{1})
    - \widehat{\mathcal{V}}(\mathbf{0}).
  \label{eq:sample-analogues}
\end{equation}

\subsection{Influence function of the WLS Shapley estimator}\label{sec:IF}

The preliminary estimator $\widetilde{\boldsymbol\phi}$ depends on $p_{\mathcal{UC}}$ only through $\mathbf{b}$ and~$c$ in~\eqref{eq:Sigma-b-c}; the matrix~$\Sigma$ is determined by the Shapley kernel alone. Replacing the population expectation by the $\mathcal{LC}$ sample average perturbs $\mathbf{b}$ and~$c$, and the resulting perturbation of~$\bar{\boldsymbol\phi}$ can be expressed in terms of per-observation contributions via the chain rule.

For each observation $(\mathbf{x}, y)$ and coalition $\mathbf{z}$, let
\begin{equation}
  \epsilon(\mathbf{z}, \mathbf{x}, y)
  \;=\;
  \eta(\mathbf{z}, \mathbf{x}, y)
  - \mathcal{V}(\mathbf{z})
  \label{eq:residual}
\end{equation}
denote the residual of the per-observation value contribution about its population mean, and define
\begin{equation}
  \mathbf{g}(\mathbf{x}, y)
  \;=\;
  \mathbb{E}_{\mu_{\mathrm{Sh}}}
  \!\big[\mathbf{z}\,
    \epsilon(\mathbf{z}, \mathbf{x}, y)\big].
  \label{eq:g-def}
\end{equation}
The influence function of the WLS Shapley estimator~\eqref{eq:kkt} is (the full derivation is given in Appendix~\ref{app:IF-derivation})
\begin{equation}
  \bar{\boldsymbol\psi}(\mathbf{x}, y)
  \;=\;
  \Sigma^{-1}
  \big[\mathbf{g}(\mathbf{x}, y)
       + \lambda(\mathbf{x}, y)\,\mathbf{1}\big],
  \label{eq:IF}
\end{equation}
where
\begin{equation}
  \lambda(\mathbf{x}, y)
  \;=\;
  \frac{
    \big(\eta(\mathbf{1}, \mathbf{x}, y)
    - \eta(\mathbf{0}, \mathbf{x}, y)\big)
    - c
    - \mathbf{1}^\top\Sigma^{-1}\mathbf{g}(\mathbf{x}, y)
  }{
    \mathbf{1}^\top\Sigma^{-1}\mathbf{1}
  }
  \label{eq:lambda}
\end{equation}
is the Lagrange correction enforcing the efficiency constraint at the observation level. 
Since the closed-form solution~\eqref{eq:kkt} is linear in $(\mathbf{b}, c)$ with $\Sigma$ fixed, the estimation error decomposes as
\begin{equation}
  \widetilde{\boldsymbol\phi}
  - \bar{\boldsymbol\phi}
  \;=\;
  \frac{1}{n}\sum_{i \in \mathcal{LC}}
  \bar{\boldsymbol\psi}(\mathbf{x}_i, y_i)
  \;+\; O_p(m^{-1/2}),
  \label{eq:IF-linearization}
\end{equation}
where $\bar{\boldsymbol\psi}$ is given by~\eqref{eq:IF}--\eqref{eq:lambda} and the remainder arises from approximating $\Sigma$ and $\mathbf{b}$ with $m$ sampled coalitions. 
The leading term is exact in the data-sampling component, as no higher-order remainder in $n$ is incurred.

\begin{Remark}[Extension to covariate shift]
\label{rem:shift}
The influence function in~\eqref{eq:IF}--\eqref{eq:lambda} is derived under
$p_{\mathcal{LC}} = p_{\mathcal{UC}}$. Suppose instead that the target covariate distribution is absolutely continuous with respect to the $\mathcal{LC}$ distribution, with density ratio
$w(\mathbf{x}) = p_{\mathcal{UC}}(\mathbf{x}) \,/\, p_{\mathcal{LC}}(\mathbf{x})$.
When $w$ is known, the same derivation applies after replacing $\eta(\mathbf{z}, \mathbf{x}, y)$ by
$w(\mathbf{x})\,\eta(\mathbf{z}, \mathbf{x}, y)$.
In particular, the weighted residual is $\epsilon_w(\mathbf{z}, \mathbf{x}, y) = w(\mathbf{x})\,\eta(\mathbf{z}, \mathbf{x}, y) - \mathcal{V}(\mathbf{z})$, and $\mathbf{g}$,
$\lambda$, and $\bar{\boldsymbol\psi}$ are defined analogously using $\epsilon_w$.

In practice, $w$ is replaced by an estimate $\widehat{w}$ obtained from the $\mathcal{LC}$ and
$\mathcal{UC}$ covariates. The linearization in~\eqref{eq:IF-linearization} continues to hold with
the same first-order influence function whenever the contribution from density-ratio estimation is
asymptotically negligible. A sufficient condition is $\|\widehat{w} - w\|_\infty = o_p(n^{-1/2})$,
together with appropriate moment and regularity conditions on~$\eta$. We treat this condition as an assumption in the present analysis; more generally, when density-ratio estimation contributes at first order, an orthogonal/debiased construction is required to account for this additional nuisance-estimation error.
\end{Remark}

\subsection{Variance reduction via partially-observed sites}
The decomposition~\eqref{eq:IF-linearization} reveals the structure that enables variance reduction. 
The estimation error is a sample average of per-observation contributions $\bar{\boldsymbol\psi}(\mathbf{x}_i, y_i)$ whose population mean is zero: $\mathbb{E}_{p_{\mathcal{UC}}} [\bar{\boldsymbol\psi}(\mathbf{X}, Y)] = \mathbf{0}$.
Each $\bar{\boldsymbol\psi}(\mathbf{x}, y)$ depends on the full feature vector~$\mathbf{x}$ and the outcome~$y$, both partially available at $\mathcal{LM}_r$, which observes $(\mathbf{X}_{\Gamma_r}, Y)$. 

The component of $\bar{\boldsymbol\psi}$ predictable from these observed variables is the conditional expectation
\begin{equation}
  \boldsymbol\tau_r^*(\mathbf{X}_{\Gamma_r}, Y)
  \;=\;
  \mathbb{E}\big[
    \bar{\boldsymbol\psi}(\mathbf{X}, Y)
    \;\big|\;
    \mathbf{X}_{\Gamma_r},\, Y
  \big].
  \label{eq:ideal-cv}
\end{equation}
In practice $\boldsymbol\tau_r^*$ is unknown. 
We estimate it from the $\mathcal{LC}$ sample by regressing the estimated influence function $\widetilde{\boldsymbol\psi}(\mathbf{x}_i, y_i)$ on $(\mathbf{X}_{\Gamma_r}, Y)$ via cross-fitted ridge regression, yielding $\widehat{\boldsymbol\tau}_r (\mathbf{X}_{\Gamma_r}, Y)$.

\medskip
\noindent\textbf{The augmented estimator.}
The preliminary estimate is corrected by adding, for each site, the difference between the $\mathcal{LM}_r$ and $\mathcal{LC}$ averages of the estimated control variate:
\begin{equation}
  \widehat{\boldsymbol\phi}_{\mathrm{aug}}
  \;=\;
  \widetilde{\boldsymbol\phi}
  \;+\;
  \sum_{r=1}^{R}
  \bigg\{
      \frac{1}{n_r}\!\sum_{(\mathbf{x},y)
        \in \mathcal{LM}_r}
      \! \widehat{\boldsymbol\tau}_r
        (\mathbf{x}_{\Gamma_r}, y)
    \;-\;
      \frac{1}{n}\!\sum_{(\mathbf{x},y)
        \in \mathcal{LC}}
      \!\widehat{\boldsymbol\tau}_r
        (\mathbf{x}_{\Gamma_r}, y)
  \bigg\}.
  \label{eq:augmented-simple}
\end{equation}
When $p_{\mathcal{LC}} = p_{\mathcal{LM}_r} = p_{\mathcal{UC}}$, each correction term has population mean zero and the augmentation reduces variance without introducing bias.
Under covariate shift, both averages require importance weighting to the target distribution~$p_{\mathcal{UC}}$; the generalization is given in~\eqref{eq:augmented-calib} of Section~\ref{sec:calibration}.

\subsection{Screening for misaligned sites}
\label{sec:screening}
The augmented estimator~\eqref{eq:augmented-simple} benefits from site~$r$ only if the correction term $\widehat{\boldsymbol\tau}_r$ estimated on $\mathcal{LM}_r$ is consistent with the same quantity
estimated on $\mathcal{LC}$. 
When the two disagree systematically, whether due to distributional incompatibility between site~$r$ and the target population or because the regression $\widehat{\boldsymbol\tau}_r$ extrapolates poorly on $\mathcal{LM}_r$ data, including site~$r$ degrades rather than improves the estimate.

For each site~$r$, we compare the importance-weighted averages of $\widehat{\boldsymbol\tau}_r$ computed on $\mathcal{LM}_r$ and $\mathcal{LC}$. Define
\begin{equation}
  \bar{\boldsymbol\tau}_r^{\,\mathcal{LM}}
  = \frac{1}{n_r}\sum_{(\mathbf{x},y)
    \in \mathcal{LM}_r}
  \widehat{w}_r(\mathbf{x}_{\Gamma_r})\,
  \widehat{\boldsymbol\tau}_r
    (\mathbf{x}_{\Gamma_r}, y),
  \qquad
  \bar{\boldsymbol\tau}_r^{\,\mathcal{LC}}
  = \frac{1}{n}\sum_{(\mathbf{x},y)
    \in \mathcal{LC}}
  \widehat{w}(\mathbf{x})\,
  \widehat{\boldsymbol\tau}_r
    (\mathbf{x}_{\Gamma_r}, y),
  \label{eq:screening}
\end{equation}
where $\widehat{w}_r(\mathbf{x}_{\Gamma_r})
= \widehat{p}_{\mathcal{UC}}(\mathbf{x}_{\Gamma_r})
  \,/\,
  \widehat{p}_{\mathcal{LM}_r}(\mathbf{x}_{\Gamma_r})$
and $\widehat{w}(\mathbf{x})
= \widehat{p}_{\mathcal{UC}}(\mathbf{x})
  \,/\, \widehat{p}_{\mathcal{LC}}(\mathbf{x})$
are the density ratios. Both reweight to~$p_{\mathcal{UC}}$, so under Assumption~\ref{asm:mar} their difference
$\mathbf{d}_r =
\bar{\boldsymbol\tau}_r^{\,\mathcal{LM}} -
\bar{\boldsymbol\tau}_r^{\,\mathcal{LC}}
\in \mathbb{R}^p$ has population mean zero. 
To ensure that coordinates with noisier importance weights do not dominate the comparison, we studentize the difference. The test statistic is
\begin{equation}
  T_r
  \;=\;
  \sum_{j=1}^{p}
  \frac{d_{r,j}^2}
       {\widehat{\mathrm{Var}}(d_{r,j})},
  \label{eq:screening-stat}
\end{equation}
where
$\widehat{\mathrm{Var}}(d_{r,j})
= \widehat{\mathrm{Var}}_{\mathcal{LC}}
  (\widehat{w}\,\widehat{\tau}_{r,j}) \,/\, n
+ \widehat{\mathrm{Var}}_{\mathcal{LM}_r}
  (\widehat{w}_r\,\widehat{\tau}_{r,j}) \,/\, n_r$
is the estimated variance of the $j$-th coordinate of the weighted mean difference.

Since the null distribution of~$T_r$ depends on the estimated importance weights and control variates in a complex way, we assess significance via a permutation test rather than a $\chi^2_p$ approximation. 
Let $\boldsymbol\tau_i^{\mathcal{LC}} = \widehat{w}(\mathbf{x}_i)\, \widehat{\boldsymbol\tau}_r(\mathbf{x}_{i,\Gamma_r}, y_i)$ denote the weighted control variate value for $\mathcal{LC}$ observation~$i$, and define $\boldsymbol\tau_k^{\mathcal{LM}}$ analogously for $\mathcal{LM}_r$ observation~$k$. 
The permutation procedure is:
\begin{enumerate}
\item Pool
  $\{\boldsymbol\tau_i^{\mathcal{LC}}\}_{i=1}^{n}$
  and
  $\{\boldsymbol\tau_k^{\mathcal{LM}}\}_{k=1}^{n_r}$
  into a combined set of $n + n_r$ observations.
\item For $b = 1, \ldots, B_{\mathrm{perm}}$: randomly
  assign $n$ observations to the $\mathcal{LC}$ group and $n_r$ to the $\mathcal{LM}$ group; compute the permuted test statistic $T_r^{(b)}$ as
  in~\eqref{eq:screening-stat}.
\item The $p$-value is
  $\hat{p}_r = (1 + \sum_{b=1}^{B_{\mathrm{perm}}}
  \mathbf{1}\{T_r^{(b)} \geq T_r\})
  \,/\, (1 + B_{\mathrm{perm}})$.
\end{enumerate}
We use $B_{\mathrm{perm}} = 1{,}000$ throughout. 
This approach avoids parametric distributional assumptions on the test statistic. While exact exchangeability under estimated nuisance parameters is not formally guaranteed, we empirically assess the calibration and power of the resulting screening procedure under both aligned and misaligned sources in Section~\ref{sec:exp3}.
Sites with $\hat{p}_r < \alpha$ are excluded from the summation in~\eqref{eq:augmented-calib}.

\subsection{Calibration}
\label{sec:calibration}
The augmented estimator~\eqref{eq:augmented-simple} applies unit weight to each site's correction. In practice, the optimal weight should depend on how well $\widehat{\boldsymbol\tau}_r$ predicts $\bar{\boldsymbol\psi}$ at site~$r$. We therefore introduce a scalar calibration weight $\delta_r$ for each site:
\begin{equation}
  \widehat{\boldsymbol\phi}_{\mathrm{aug}}
  \;=\;
  \widetilde{\boldsymbol\phi}
  \;+\;
  \sum_{r=1}^{R}
  \delta_r
  \bigg\{
    \frac{1}{n_r}\!\sum_{(\mathbf{x},y)
      \in \mathcal{LM}_r}
    \!\widehat{\boldsymbol\tau}_r
      (\mathbf{x}_{\Gamma_r}, y)
    \;-\;
    \frac{1}{n}\!\sum_{(\mathbf{x},y)
      \in \mathcal{LC}}
    \!\widehat{\boldsymbol\tau}_r
      (\mathbf{x}_{\Gamma_r}, y)
  \bigg\}.
  \label{eq:augmented-calib}
\end{equation}
The weight $\delta_r$ is chosen to minimize the empirical variance of $\widehat{\boldsymbol\phi}_{\mathrm{aug}}$. 
For each feature~$j$, this reduces to a quadratic program in $R$ variables with closed-form solution $\delta_{r,j}^* = [A_j^{-1}\mathbf{b}_j]_r$ (derived in Appendix~\ref{app:calibration}), where
\begin{align}
  [A_j]_{rs}
  &= \widehat{\mathrm{Cov}}_{\mathcal{LC}}
     (\widehat\tau_{r,j},\;
      \widehat\tau_{s,j})
     \;+\;
     \mathbf{1}_{r=s}\,
       \frac{n}{n_r}\,
       \widehat{\mathrm{Var}}_{\mathcal{LM}_r}
       (\widehat\tau_{r,j}),
  \label{eq:calib-A}
  \\[4pt]
  [\mathbf{b}_j]_r
  &= \widehat{\mathrm{Cov}}_{\mathcal{LC}}
     (\widetilde\psi_j,\;
      \widehat\tau_{r,j}).
  \label{eq:calib-b}
\end{align}
The per-site weight minimizes the total variance $\sum_{j=1}^{p}\mathrm{Var}(\widehat\phi_{\mathrm{aug},j})$ directly:
\begin{equation}
  \boldsymbol\delta^*
  \;=\;
  \bigg(\sum_{j=1}^{p} A_j\bigg)^{-1}
  \bigg(\sum_{j=1}^{p} \mathbf{b}_j\bigg),
  \label{eq:delta-total}
\end{equation}
a single $R$-dimensional linear system obtained by summing the per-feature quadratic objectives.

The control variate $\widehat{\boldsymbol\tau}_r$ is trained via cross-fitting on $\mathcal{LC}$. The complete procedure is summarized in Algorithm~\ref{alg:fushap}.

\section{Simulations}
\label{sec.simulations}
We compare FUSHAP against six types of baselines that represent the principal strategies available when labeled data are distributed across sites with blockwise-missing features. 
All methods target the same estimand (the global Shapley attribution vector $\bar{\boldsymbol\phi}$ on the $\mathcal{UC}$ population) and share the same $\mathcal{UC}$ background sample for the restricted prediction~\eqref{eq:restricted-pred}. They differ only in which labeled observations are used to estimate the value function $\mathcal{V}(\mathbf{z})$. The same baselines are used in the real-data applications of Section~\ref{sec.realdat}.

\begin{itemize}
\item[\textbf{(A)}] \textbf{Single-site estimator.} The WLS Shapley estimator applied to $\mathcal{LC}$ alone, with $\mathcal{UC}$ as the background sample for marginalizing absent features. Unbiased but potentially high-variance due to the small $\mathcal{LC}$ sample.

\item[\textbf{(B)}] \textbf{Single-site with importance weighting.} Identical to~(A) but with each $\mathcal{LC}$ observation reweighted by the estimated density ratio $\hat{w}(x) = \hat{p}_{\mathcal{UC}}(x) / \hat{p}_{\mathcal{LC}}(x)$ to correct for covariate shift. Equivalent to FUSHAP with all calibration weights set to zero; serves as a direct ablation.

\item[\textbf{(C)}] \textbf{Impute-then-estimate.}
Missing features at each $\mathcal{LM}_r$ are imputed from $\mathcal{LC}$ reference values and the completed data are pooled with $\mathcal{LC}$.
Since prior work and our simulations indicate that switching among standard imputers does not necessarily resolve attribution bias under blockwise missingness~\citep{vo2025explainabilitymachinelearningmodels}, we report two representative methods: mean imputation and MICE (iterative conditional imputation).
  
\item[\textbf{(D)}] \textbf{Per-site averaging.} Each site independently imputes, estimates $\widehat{\mathcal{V}}_r(\mathbf{z})$ from its own labeled observations, solves the WLS, and the resulting Shapley vectors are averaged weighted by sample size.

\item[\textbf{(E)}] \textbf{Complete-case.} Only features observed at every site ($\bigcap_r \Gamma_r$) are retained; $\eta$ is averaged over all labeled observations using this reduced feature set.

\item[\textbf{(F)}] \textbf{Oracle.} All $\mathcal{LC}$ and $\mathcal{LM}_r$ observations are pooled with the missing features at each $\mathcal{LM}_r$ site treated as observed, yielding $n + \sum_r n_r$ labeled observations with complete feature vectors on which the WLS Shapley estimator is applied.

\end{itemize}

All baselines compute $\hat{\boldsymbol\phi}$ via the WLS characterization~\eqref{eq:wls} with $m$ sampled coalitions from $\mu_{\mathrm{Sh}}$. 
For each simulation configuration, the reference $\bar{\boldsymbol\phi}^{\mathrm{true}}$ is a high-precision Monte Carlo approximation to the population Shapley vector, computed by exact enumeration over all $2^p$ coalitions using the combinatorial Shapley formula, with each value function $\mathcal{V}(S)$ evaluated as the sample average of $\eta(\mathbf{z}_S, \mathbf{x}_i, y_i)$ over a large independent evaluation set ($n_{\mathrm{eval}} = 50{,}000$ observations from $p_{\mathcal{UC}}$, with $K = 500$ background samples for the marginal imputation).
This evaluation set is generated independently of the $\mathcal{UC}$ sample used by the methods. 
The reference is computed once per configuration and held fixed across all $B$ replications; therefore differences in MSE across methods reflect the estimation strategy rather than variation in the evaluation target.

\subsection{Simulation Setup}
\subsubsection{Data-generating process}
\label{sec.dgp}

The target population $\mathcal{UC}$ has features
drawn as $\mathbf{X} \sim N(\mathbf{0}, I_p)$ with
$p = 10$.
The outcome is generated as $Y = f(\mathbf{X}) + \varepsilon$ with $\varepsilon \sim N(0, 0.25)$, independently of~$\mathbf{X}$. 
We consider three outcome models of increasing complexity.

\paragraph{Model~I (linear).}
\begin{equation}
  f_{\mathrm{I}}(\mathbf{x})
  \;=\;
  \sum_{j=1}^{p} \beta_j\, x_j,
  \qquad
  \beta_j = \frac{p + 1 - j}{p},
  \label{eq:model-linear}
\end{equation}
a linear predictor with monotonically decreasing coefficients. Under squared-error loss, the per-observation value contribution $\eta(\mathbf{z}, \mathbf{x}, y)$ is quadratic in~$(\mathbf{x}, y)$, and the influence function $\bar{\boldsymbol\psi}$ inherits this polynomial structure.

\paragraph{Model~II (polynomial interactions).}
\begin{equation}
  f_{\mathrm{II}}(\mathbf{x})
  \;=\;
  f_{\mathrm{I}}(\mathbf{x})
  + 0.5\,x_1 x_2 + 0.3\,x_3 x_4 + 0.4\,x_5^2,
  \label{eq:model-poly}
\end{equation}
augmenting Model~I with pairwise interactions and a quadratic term. The influence function retains polynomial dependence on the data, though of higher degree than in Model~I.

\paragraph{Model~III (non-polynomial).}
\begin{equation}
  f_{\mathrm{III}}(\mathbf{x})
  \;=\;
  \sin(x_1 x_2) + 0.8\max(x_3 + x_4,\, 0)
  + 0.5\,x_5^2 - 0.3\,x_6,
  \label{eq:model-mixed}
\end{equation}
a non-polynomial model whose influence function cannot be fully captured by polynomial control variates.

\subsubsection{Multi-site data construction}
From each model, we construct the multi-site data structure of Section~\ref{sec.data} by drawing $n + \sum_r n_r + N$ independent observations and allocating them to $\mathcal{LC}$ ($n = 300$), $R = 3$ labeled-missing sources ($n_r = 2{,}000$ each), and $\mathcal{UC}$ ($N = 5{,}000$). 
The missing-feature blocks are non-overlapping: site~$r$ observes all features except $\{2(r{-}1){+}1,\, 2r\}$, so that each site lacks $20\%$ of the feature set and $\mathcal{LC}$ is the only source with complete coverage.

To reflect the heterogeneity typical of multi-site studies, each data source is subject to both location and scale shifts: features at source~$s$ are drawn as $\mathbf{X}^{(s)} \sim N(\delta_s \mathbf{1},\, \sigma_s^2 I_p)$, 
with $(\delta, \sigma) = (0, 1)$ for $\mathcal{UC}$, $(0.1, 1.05)$ for $\mathcal{LC}$, $(-0.15, 0.9)$ for $\mathcal{LM}_1$, $(0.2, 1.1)$ for $\mathcal{LM}_2$, and $(-0.1, 0.95)$ for $\mathcal{LM}_3$. 
The density ratio $\widehat{w}(\mathbf{x}) = \widehat{p}_{\mathcal{UC}}(\mathbf{x}) / \widehat{p}_s(\mathbf{x})$ is estimated via a gradient-boosted classifier on the combined sample with source indicators.
In Experiments~1, 2, and~4, the location-scale shift is applied to all $p$ features at each site, including those subsequently declared missing. Consequently, these experiments deliberately introduce moderate violations of Assumption~\ref{asm:mar} and evaluate FUSHAP beyond its exact alignment regime. In Experiment 3, where screening calibration and power are the quantities of interest, aligned sites are instead constructed to satisfy Assumption 1 exactly (the shift is applied only to the observed features $\mathbf{X}_{\Gamma_r}$, while the missing features $\mathbf{X}_{\Gamma_r^c}$ are drawn from the target distribution $p_{\mathcal{UC}}$).

\subsection{Experiments}
\label{sec:experiments}

All methods receive the same fixed, pre-trained predictive model and compute its Shapley attribution vector. 
In the simulation studies, this is the true data-generating function~$f$; in the real-data applications (Section~\ref{sec.realdat}), this is a model trained on held-out data. 
Across $B$ independent replications, we report the mean squared error
$\mathrm{MSE} = \mathrm{Bias}^2 + \mathrm{Var}$,
where $\mathrm{Bias}^2 =
\|\bar{\hat{\boldsymbol\phi}}
- \bar{\boldsymbol\phi}\|^2$ and
$\mathrm{Var} = \sum_j \widehat{\mathrm{Var}}_B(\hat\phi_j)$, together with Spearman's rank correlation $\rho_s$ between each estimate and the ground truth to assess agreement in the feature importance ranking. 
Where appropriate, we report the variance ratio $\mathrm{VR} = \mathrm{Var}(\text{method})\,/\, \mathrm{Var}(\text{LC-only})$.

\begin{table}[!htbp]
\centering
\caption{Experiment~1: MSE decomposition and Spearman
rank correlation ($B = 100$, $p = 10$, $n = 300$,
$n_r = 2{,}000$, $R = 3$). Best feasible method in
bold.}
\label{tab:exp1}
\small

\medskip
\textit{Model~I (linear)}
\medskip

\begin{tabular}{l cccc}
\toprule
Method & MSE & Bias$^2$ & Var & $\rho_s$ \\
\midrule
(F) Oracle       & 0.032 & 0.027 & 0.004 & 1.00 \\
\textbf{FUSHAP}  & \textbf{0.052} & 0.003 & 0.049 & 0.98 \\
(A) Single-site  & 0.175 & 0.087 & 0.088 & 0.98 \\
(B) Single+IPW   & 0.092 & 0.019 & 0.074 & 0.98 \\
(C) Impute-mean  & 0.118 & 0.111 & 0.008 & 0.99 \\
(C) Impute-MICE  & 0.120 & 0.112 & 0.008 & 0.99 \\
(D) Per-site avg & 0.119 & 0.111 & 0.009 & 0.99 \\
(E) Complete-case& 2.492 & 2.491 & 0.001 & $-$0.75 \\
\bottomrule
\end{tabular}

\medskip
\textit{Model~II (sparse interactions)}
\medskip

\begin{tabular}{l cccc}
\toprule
Method & MSE & Bias$^2$ & Var & $\rho_s$ \\
\midrule
(F) Oracle       & 0.065 & 0.059 & 0.006 & 0.99 \\
\textbf{FUSHAP}  & \textbf{0.065} & 0.005 & 0.060 & 0.98 \\
(A) Single-site  & 0.550 & 0.367 & 0.183 & 0.97 \\
(B) Single+IPW   & 0.109 & 0.015 & 0.093 & 0.97 \\
(C) Impute-mean  & 0.129 & 0.118 & 0.011 & 0.98 \\
(C) Impute-MICE  & 0.130 & 0.119 & 0.011 & 0.98 \\
(D) Per-site avg & 0.130 & 0.118 & 0.013 & 0.98 \\
(E) Complete-case& 3.336 & 3.334 & 0.002 & $-$0.75 \\
\bottomrule
\end{tabular}

\medskip
\textit{Model~III (non-polynomial)}
\medskip

\begin{tabular}{l cccc}
\toprule
Method & MSE & Bias$^2$ & Var & $\rho_s$ \\
\midrule
(F) Oracle       & 0.002 & 0.001 & 0.001 & 0.96 \\
\textbf{FUSHAP}  & \textbf{0.020} & 0.009 & 0.010 & 0.94 \\
(A) Single-site  & 0.061 & 0.036 & 0.025 & 0.95 \\
(B) Single+IPW   & 0.025 & 0.015 & 0.011 & 0.95 \\
(C) Impute-mean  & 0.062 & 0.060 & 0.002 & 0.86 \\
(C) Impute-MICE  & 0.062 & 0.060 & 0.002 & 0.86 \\
(D) Per-site avg & 0.062 & 0.060 & 0.002 & 0.86 \\
(E) Complete-case& 0.374 & 0.374 & 0.000 & --- \\
\bottomrule
\end{tabular}
\end{table}

\subsubsection{Experiment~1 (imputation bias).}\label{sec:exp1} 
Under the default configuration with all three outcome models, we compare all baselines and FUSHAP over $B = 100$ replications. 
Table~\ref{tab:exp1} reports the results. 
FUSHAP achieves the lowest MSE among all feasible methods across all three models, with improvements of $3.4\times$ (Model~I), $8.5\times$ (Model~II), and $3.1\times$ (Model~III) over the single-site estimator.
The gains are largest for Models~I and~II, where the polynomial control variate is well-specified and captures a substantial fraction of the influence function's variability. 
Under Model~III (non-polynomial), the control variate approximation is less effective, yet FUSHAP still achieves $3.1\times$ lower MSE than imputation.

The bias--variance decomposition reveals the mechanism.
The single-site estimator has low bias but high variance ($0.088$ in Model~I); imputation baselines reduce variance ($0.008$) but introduce substantial bias ($0.111$). 
FUSHAP achieves both low bias ($0.003$) and moderate variance ($0.049$), outperforming all alternatives in total MSE. 
Notably, mean imputation and MICE produce nearly identical results, suggesting that switching between these standard imputation procedures alone does not eliminate the attribution bias~\citep{vo2025explainabilitymachinelearningmodels}.

\subsubsection{Experiment~2 (variance reduction).}
\label{sec:exp2}

\begin{figure*}[!htbp]
\centering
\begin{subfigure}[b]{\textwidth}
  \centering
  \includegraphics[width=\textwidth]{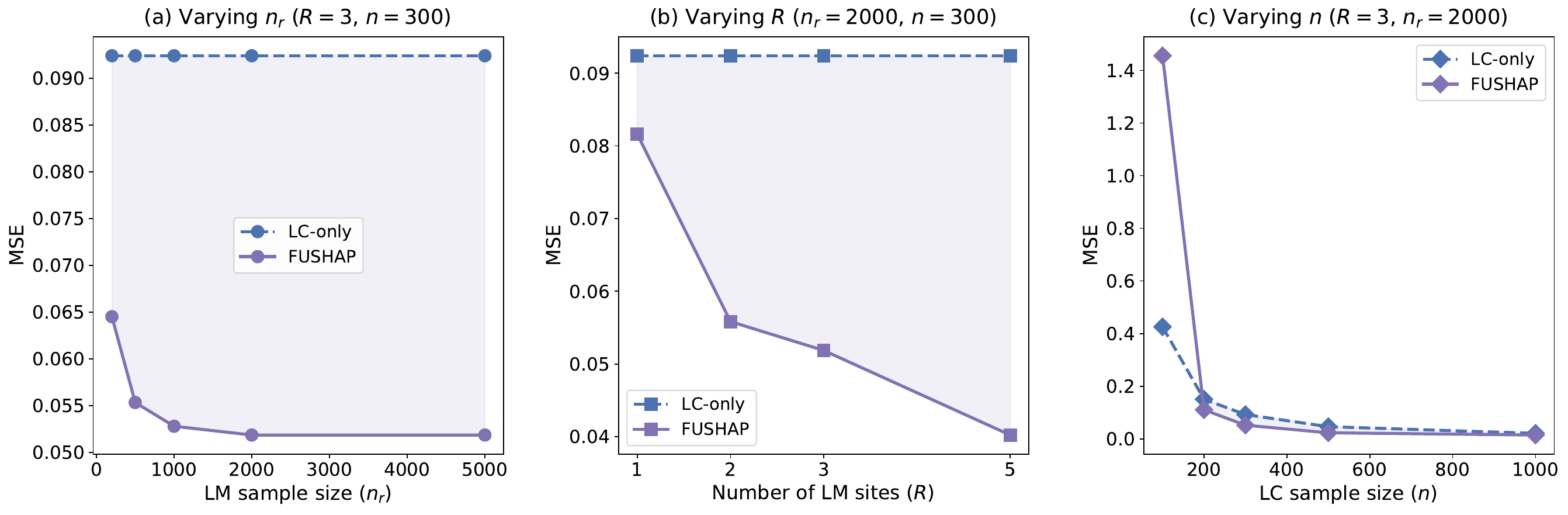}
  \caption{Model~I (linear)}
  \label{fig:exp2-I}
\end{subfigure}\\[6pt]
\begin{subfigure}[b]{\textwidth}
  \centering
  \includegraphics[width=\textwidth]{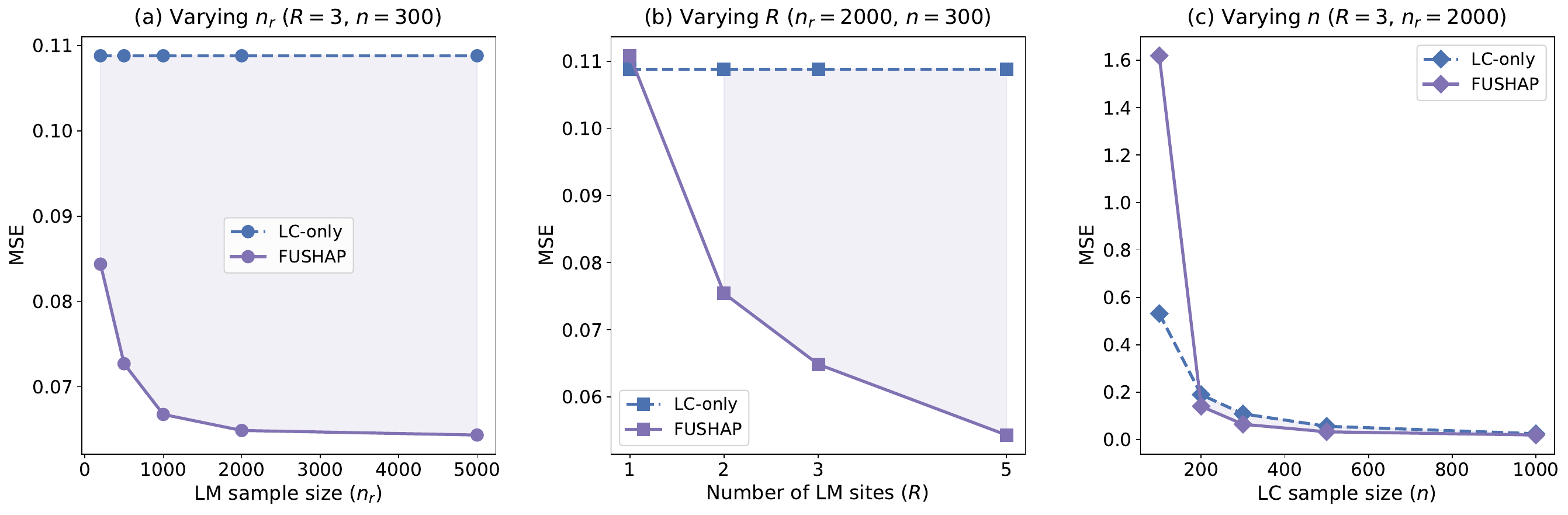}
  \caption{Model~II (sparse interactions)}
  \label{fig:exp2-II}
\end{subfigure}\\[6pt]
\begin{subfigure}[b]{\textwidth}
  \centering
  \includegraphics[width=\textwidth]{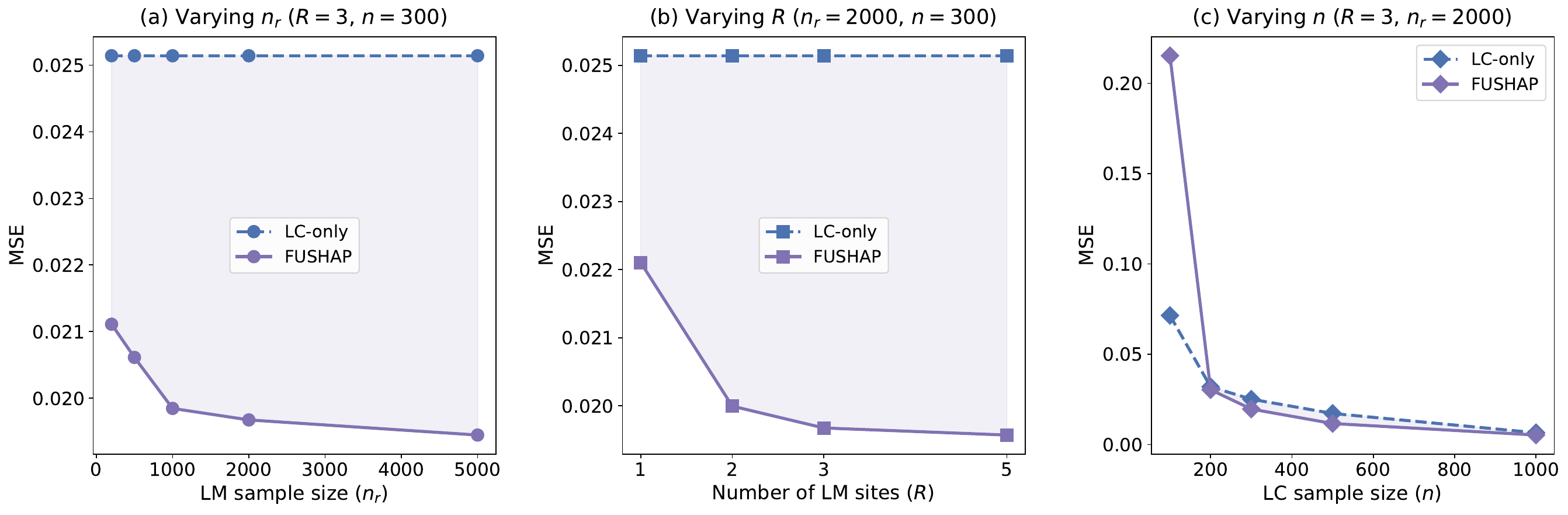}
  \caption{Model~III (non-polynomial)}
  \label{fig:exp2-III}
\end{subfigure}
\caption{Experiment~2: MSE of FUSHAP (solid) and
LC-only (dashed) as a function of $n_r$ (left),
$R$ (center), and $n$ (right), $B = 100$.}
\label{fig:exp2}
\end{figure*}

We examine how FUSHAP's MSE depends on three design parameters: the $\mathcal{LM}$ sample size~$n_r \in \{200, 500, 1{,}000, 2{,}000, 5{,}000\}$ with $R = 3$ fixed (left panels), the number of sites~$R \in \{1, 2, 3, 5\}$ with $n_r = 2{,}000$ fixed (center panels), and the $\mathcal{LC}$ sample size~$n \in \{100, 200, 300, 500, 1{,}000\}$ with $R = 3$ and $n_r = 2{,}000$ fixed (right panels).
Figure~\ref{fig:exp2} reports results under all three outcome models.

Three patterns are consistent across models. 
First, FUSHAP's improvement increases with $n_r$ but exhibits diminishing returns beyond $n_r \approx 1{,}000$ (left panels). 
Second, adding sites monotonically reduces MSE: at $R = 5$, FUSHAP achieves $2.3\times$ lower MSE than the single-site estimator under Model~I (center panels). Third, the $\mathcal{LC}$ sample size has a critical lower bound: at $n = 100$, the control variate regression overfits and FUSHAP degrades; for $n \geq 200$, FUSHAP consistently improves upon LC-only (right panels). 
The variance reduction is largest under Model~I, where the polynomial control variate is well-specified and smallest under Model~III, where the approximation is less effective.
FUSHAP requires a sufficient number of complete observations to learn the influence-function projection reliably; when the $\mathcal{LC}$ sample is very small ($n = 100$), the control variate regression overfits and auxiliary data cannot compensate.

\subsubsection{Experiment~3 (screening).}
\label{sec:exp3}

Under Model~I with $R = 4$ sites, the first three sites satisfy Assumption~\ref{asm:mar} exactly: the covariate shift is applied only to the observed features $\mathbf{X}_{\Gamma_r}$, while the missing features $\mathbf{X}_{\Gamma_r^c}$ are drawn from $p_{\mathcal{UC}}$. 
The fourth site is misaligned: its outcome is generated under perturbed coefficients $\tilde{\beta}_j = \beta_j + \Delta$ for $j \in \Gamma_4^c$.

\begin{figure}[!htbp]
\centering
\includegraphics[width=\textwidth]{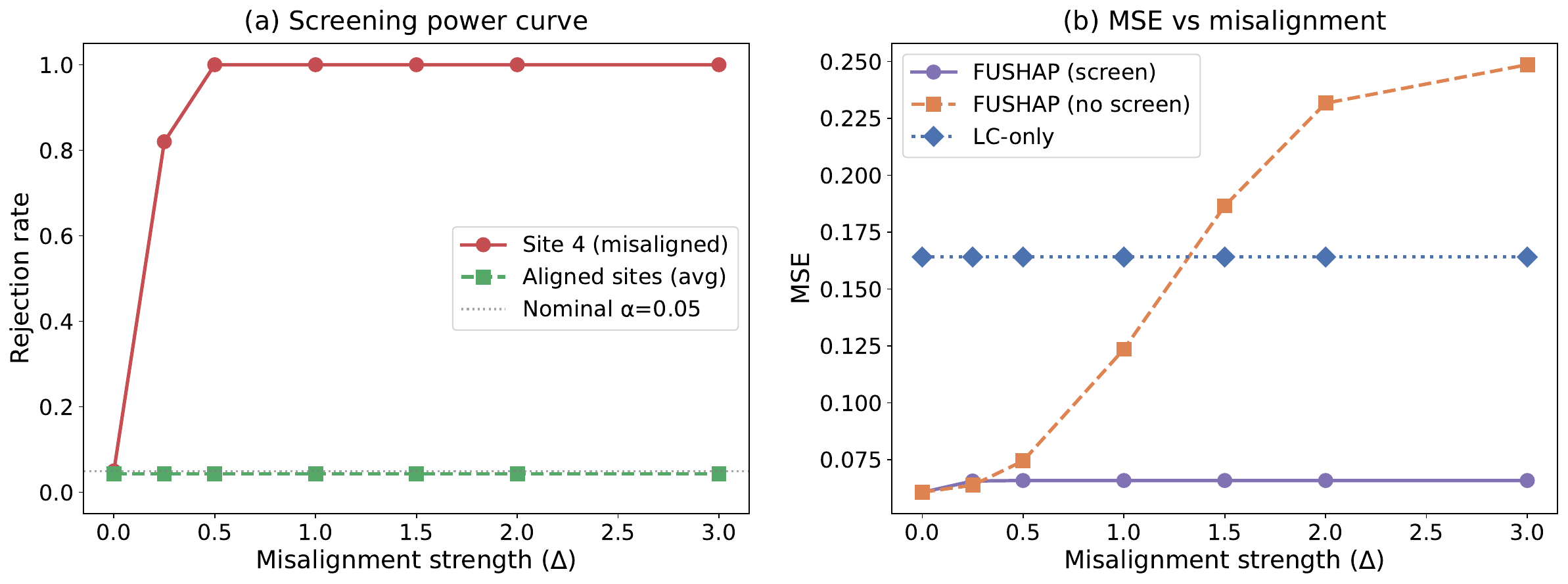}
\caption{Experiment~3: screening power and MSE
($B = 100$, $\alpha = 0.05$). Left: rejection rate
of the misaligned site and aligned sites (average).
Right: MSE of FUSHAP with and without screening.}
\label{fig:exp3}
\end{figure}

Figure~\ref{fig:exp3} reports the rejection rate (left) and MSE (right) as $\Delta$ varies from $0$ to $3$ over $B = 100$ replications.
At $\Delta = 0$ (no misalignment), the average
rejection rate of the three aligned sites is $4.3\%$, close to the nominal $\alpha = 0.05$ (Table~\ref{tab:exp3-detail}). The misaligned site is detected with $82\%$ power at $\Delta = 0.25$ and $100\%$ for $\Delta \geq 0.5$.
Without screening, FUSHAP's MSE degrades from $0.061$ at $\Delta = 0$ to $0.249$ at $\Delta = 3$, exceeding the single-site baseline ($0.164$). 
With screening, MSE stabilizes between $0.061$ and $0.066$ across all values of $\Delta$, a roughly $2.5\times$ improvement over the single-site estimator.
Detailed per-site results are reported in Table~\ref{tab:exp3-detail} of Appendix~\ref{app:additional}.

\subsubsection{Experiment~4 (computational cost).}
\label{sec:exp4}

Table~\ref{tab:exp4} reports wall-clock time as the number of features increases. 
FUSHAP's overhead relative to the single-site estimator is modest ($3$--$5\times$) and arises from the control variate regression and calibration steps. 
Compared to the impute-then-pool baseline, FUSHAP is $2.7$--$4.6\times$ faster because it computes Shapley values on the small $\mathcal{LC}$ sample ($n = 300$) rather than the pooled dataset ($n + \sum_r n_r = 6{,}300$). 
The speedup grows with $p$ because the pooled Shapley computation scales with both sample size and the number of coalitions.

\begin{table}[!htbp]
\centering
\caption{Experiment~4: Wall-clock time in seconds
(mean $\pm$ std over 3 runs, $n = 300$,
$n_r = 2{,}000$, $R = 2$).}
\label{tab:exp4}
\small
\begin{tabular}{r ccc}
\toprule
$p$ & LC-only & Impute-pool & FUSHAP \\
\midrule
4  & $0.47 \pm 0.01$ & $6.79 \pm 0.08$
   & $2.52 \pm 0.02$ \\
6  & $0.67 \pm 0.02$ & $9.39 \pm 0.00$
   & $2.79 \pm 0.02$ \\
8  & $0.77 \pm 0.01$ & $11.65 \pm 0.07$
   & $3.04 \pm 0.04$ \\
10 & $0.99 \pm 0.01$ & $14.43 \pm 0.18$
   & $3.41 \pm 0.03$ \\
12 & $1.10 \pm 0.02$ & $16.48 \pm 0.20$
   & $3.62 \pm 0.12$ \\
\bottomrule
\end{tabular}
\end{table}

\section{Real-data Applications}
\label{sec.realdat}
We evaluate FUSHAP on two real-world datasets with controlled blockwise missingness imposed on the auxiliary sites, enabling quantitative comparison against a ground-truth attribution vector.
Since mean imputation and MICE produce nearly identical Shapley attributions under blockwise missingness, both in our simulations (Table~\ref{tab:exp1}) and in prior work~\citep{vo2025explainabilitymachinelearningmodels}, only mean imputation is reported below.

\subsection{Beijing Multi-Site Air Quality}
\label{sec:beijing}

We apply FUSHAP to the Beijing Multi-Site Air Quality dataset~\citep{zhang2017cautionary}, which records daily averages of six pollutants and five meteorological variables at $12$ monitoring stations. The task is to attribute a model for PM2.5 concentration using the remaining $p = 10$ features. One urban station (Dongsi, $n = 300$) serves as $\mathcal{LC}$; three suburban stations serve as $\mathcal{LM}$ sites, each missing a different pair of features (Table~\ref{tab:beijing}); the remaining eight stations form $\mathcal{UC}$.
Four models are trained on $15{,}646$ observations from all non-$\mathcal{LC}$ stations and held fixed during Shapley estimation.
The reference attribution vector is computed from $2{,}000$ held-out training observations with $m = 300$ sampled coalitions.

\begin{table}[!htbp]
\centering
\caption{Beijing Air Quality: data partition and
results ($\times 10^{-3}$, $B = 50$). Best feasible
method in bold.}
\label{tab:beijing}
\small

\medskip
\textit{(a) Partition}
\medskip

\begin{tabular}{l l c l}
\toprule
Source & Station & $n$ & Missing \\
\midrule
$\mathcal{LC}$ & Dongsi & 300 & --- \\
$\mathcal{LM}_1$ & Changping & 1{,}450 & CO, O3 \\
$\mathcal{LM}_2$ & Huairou & 1{,}445 & SO2, NO2 \\
$\mathcal{LM}_3$ & Shunyi & 1{,}384 & DEWP, PRES \\
$\mathcal{UC}$ & 8 stations & 11{,}375 & --- \\
\bottomrule
\end{tabular}

\bigskip
\textit{(b) MSE ($\times 10^{-3}$)}
\medskip

\begin{tabular}{l cccc}
\toprule
Method & Linear & RF & GBM & MLP \\
\midrule
(A) Single-site  & 42.7 & 42.1 & 39.4 & 39.0 \\
(B) Single+IPW   & 20.9 & 22.4 & 22.0 & 25.0 \\
(C) Impute-mean  & 12.1 & 11.8 & 12.7 & 23.3 \\
(D) Per-site avg & 11.8 & 11.6 & 12.5 & 22.8 \\
(E) Complete-case& 62.7 & 79.7 & 77.1 & 136.1 \\
\textbf{FUSHAP}
  & \textbf{5.8} & \textbf{7.3}
  & \textbf{7.4} & \textbf{8.8} \\
\bottomrule
\end{tabular}

\bigskip
\textit{(c) Bias$^2$ / Var ($\times 10^{-3}$)}
\medskip

\begin{tabular}{l cccc}
\toprule
Method
  & Linear & RF & GBM & MLP \\
\midrule
(A) Single-site
  & 13.4 / 29.2 & 13.3 / 28.7
  & 12.4 / 27.0 & 12.2 / 26.8 \\
(B) Single+IPW
  & 11.9 / 9.0 & 14.9 / 7.6
  & 14.4 / 7.6 & 14.1 / 10.9 \\
(C) Impute-mean
  & 10.7 / 1.4 & 10.6 / 1.2
  & 11.6 / 1.2 & 20.4 / 2.8 \\
(D) Per-site avg
  & 10.6 / 1.2 & 10.6 / 1.0
  & 11.5 / 1.0 & 20.7 / 2.1 \\
(E) Complete-case
  & 61.6 / 1.1 & 79.5 / 0.2
  & 76.8 / 0.2 & 134.7 / 1.5 \\
\textbf{FUSHAP}
  & \textbf{4.2 / 1.7}
  & \textbf{5.5 / 1.8}
  & \textbf{5.5 / 1.8}
  & \textbf{5.5 / 3.3} \\
\bottomrule
\end{tabular}
\end{table}

FUSHAP achieves the lowest MSE across all four models, with MSEs of $5.8$, $7.3$, $7.4$, and $8.8$ ($\times 10^{-3}$) for the linear, RF, GBM, and MLP models, respectively.
Relative to the single-site estimator, these correspond to improvements of $7.4\times$, $5.8\times$, $5.3\times$, and $4.4\times$. 
FUSHAP also improves over the best imputation-based baseline by $2.1\times$ (linear), $1.6\times$ (RF), $1.7\times$ (GBM), and $2.6\times$ (MLP). 
The bias-variance decomposition (Panel~c) shows that FUSHAP achieves both low bias and low variance across all models (e.g., Bias$^2 = 4.2$, Var $= 1.7$ under the linear model), whereas the imputation baselines trade reduced variance for substantial bias (Bias$^2 = 10.7$, Var $= 1.4$).

\subsection{NACC Alzheimer's Disease}\label{sec:nacc}

We apply FUSHAP to multi-center clinical data from the National Alzheimer's Coordinating Center (NACC), predicting Mini-Mental State Examination (MMSE) scores from $p = 9$ demographic and clinical features. 
A random sample of $n = 300$ patients serves as $\mathcal{LC}$; the remaining patients are partitioned into three $\mathcal{LM}$ sites with complementary synthetic missingness and a $\mathcal{UC}$ pool
(Table~\ref{tab:nacc}; preprocessing details in Appendix~\ref{app:nacc}). 
Complete-case analysis is infeasible because the three missingness blocks are fully complementary, leaving no features common to all sites. 
Because $\mathcal{LC}$ is a random sample from the pooled population, no covariate shift correction is needed and Baseline~(B) is omitted.
Four regression models are trained on $16{,}900$ non-$\mathcal{LC}$ patients and held fixed during Shapley estimation. 
The reference attribution vector is computed from the $16{,}900$ non-$\mathcal{LC}$ training observations with $m = 300$ sampled coalitions.

\begin{table}[!htbp]
\centering
\caption{NACC Alzheimer's: data partition and results
($B = 50$). Best feasible method in bold.}
\label{tab:nacc}
\small

\medskip
\textit{(a) Partition}
\medskip

\begin{tabular}{l c l}
\toprule
Source & $n$ & Missing \\
\midrule
$\mathcal{LC}$ (random)
  & 300 & --- \\
$\mathcal{LM}_1$
  & 3{,}756 & WEIGHT, HEIGHT, NACCLIVS \\
$\mathcal{LM}_2$
  & 3{,}756 & ALCOHOL, TOBAC100, CDRLANG \\
$\mathcal{LM}_3$
  & 3{,}755 & SEX, RACE, NACCAGE \\
$\mathcal{UC}$
  & 5{,}633 & --- \\
\bottomrule
\end{tabular}

\bigskip
\textit{(b) MSE}
\medskip

\begin{tabular}{l cccc}
\toprule
Method & Ridge & RF & GBM & MLP \\
\midrule
(A) Single-site
  & 6.83 & 8.54 & 7.35 & 7.07 \\
(B) Single+IPW
  & --- & --- & --- & --- \\
(C) Impute-mean
  & 6.23 & 16.25 & 10.67 & 21.53 \\
(D) Per-site avg
  & 6.22 & 16.26 & 10.67 & 21.59 \\
(E) Complete-case
  & N/A & N/A & N/A & N/A \\
\textbf{FUSHAP}
  & \textbf{1.03} & \textbf{3.25}
  & \textbf{2.79} & \textbf{1.79} \\
\bottomrule
\end{tabular}

\bigskip
\textit{(c) Bias$^2$ / Var}
\medskip

\begin{tabular}{l cccc}
\toprule
Method
  & Ridge & RF & GBM & MLP \\
\midrule
(A) Single-site
  & 0.80 / 6.03 & 1.83 / 6.71
  & 0.97 / 6.38 & 0.61 / 6.47 \\
(B) Single+IPW
  & --- & --- & --- & --- \\
(C) Impute-mean
  & 6.12 / 0.11 & 16.12 / 0.13
  & 10.54 / 0.13 & 21.09 / 0.45 \\
(D) Per-site avg
  & 6.11 / 0.11 & 16.13 / 0.13
  & 10.54 / 0.13 & 21.15 / 0.44 \\
(E) Complete-case
  & N/A & N/A & N/A & N/A \\
\textbf{FUSHAP}
  & \textbf{0.18 / 0.85}
  & \textbf{1.55 / 1.70}
  & \textbf{0.93 / 1.86}
  & \textbf{0.36 / 1.43} \\
\bottomrule
\end{tabular}
\end{table}

FUSHAP achieves the lowest MSE across all four models, with MSEs of $1.03$, $3.25$, $2.79$, and $1.79$ for the ridge, RF, GBM, and MLP models, respectively.
Relative to the single-site estimator, these correspond to improvements of $6.6\times$, $2.6\times$, $2.6\times$, and $4.0\times$.
FUSHAP also improves over the best imputation baseline by $6.0\times$ (ridge), $5.0\times$ (RF), $3.8\times$ (GBM), and $12.0\times$ (MLP).
For three of the four models, imputation performs worse than the single-site estimator, with MSEs up to $3.0\times$ higher (MLP), demonstrating that imputation bias can outweigh its variance reduction in this setting.
The bias-variance decomposition (Panel~c) reveals that the single-site estimator has moderate bias and high variance ($\sim\!6.5$), imputation has low variance but dominant bias ($6$--$21$), and FUSHAP achieves both low bias ($0.2$--$1.6$) and substantially reduced variance ($0.9$--$1.9$).

\section{Discussion}
\label{sec:discussion}
We proposed FUSHAP, a method for estimating global Shapley feature attributions from multi-site data with blockwise missingness. 
By deriving the influence function of the constrained WLS Shapley estimator and constructing site-specific control variates, FUSHAP reduces variance without imputing missing features, avoiding the systematic coalition-dependent bias that imputation introduces. 
A permutation-based screening procedure protects against incompatible auxiliary sites, and data-adaptive calibration weights ensure that each site's contribution is proportional to its informativeness.

Several directions provide opportunities for extending the current framework.
First, the current framework assumes that the unlabeled sample $\mathcal{UC}$ has complete feature coverage. 
Extending FUSHAP to settings where the target covariate distribution is only partially observed would broaden its applicability to more general missing-data configurations and would require additional identification assumptions.
Second, while our current implementation assumes centralized access to row-level data, the FUSHAP augmentation and calibration are constructed from
site-level averages and covariance summaries. 
This structure provides a natural starting point for privacy-preserving distributed implementations that communicate summary statistics rather than individual-level data.

\newpage
\section*{Acknowledgments}
The NACC database is funded by NIA/NIH Grant U24 AG072122. NACC data are contributed by the NIA-funded ADRCs: P30 AG062429 (PI James Brewer, MD, PhD), P30 AG066468 (PI Oscar Lopez, MD), P30 AG062421 (PI Teresa Gomez-Isla, MD), P30 AG066509 (PI Thomas Grabowski, MD), P30 AG066514 (PI Mary Sano, PhD), P30 AG066530 (PI Helena Chui, MD, Arthur Toga, PhD), P30 AG066507 (PI Marilyn Albert, PhD), P30 AG066444 (PI David Holtzman, MD), P30 AG066518 (PIs Lisa Silbert, MD, Kevin Duff, PhD), P30 AG066512 (PI Thomas Wisniewski, MD), P30 AG066462 (PI Scott Small, MD), P30 AG072979 (PI David Wolk, MD), P30 AG072972 (PIs Charles DeCarli, MD, Rachel Whitmer, PhD), P30 AG072976 (PI Andrew Saykin, PsyD), P30 AG072975 (PI Julie Schneider, MD, MS), P30 AG072978 (PI Ann McKee, MD), P30 AG072977 (PI Robert Vassar, PhD), P30 AG066519 (PI Joshua Grill, PhD), P30 AG062677 (PIs Brad Boeve, MD, Ronald Petersen, MD, PhD), P30 AG079280 (PI Jessica Langbaum, PhD), P30 AG062422 (PI Gil Rabinovici, MD), P30 AG066511 (PI Allan Levey, MD, PhD), P30 AG072946 (PI Linda Van Eldik, PhD), P30 AG062715 (PI Sanjay Asthana, MD, FRCP), P30 AG072973 (PI Russell Swerdlow, MD), P30 AG066506 (PIs Glenn Smith, PhD, ABPP, David Lowenstein, PhD, Ranjan Duara, MD), P30 AG066508 (PIs Stephen Strittmatter, MD, PhD, Christopher Van Dyck, MD), P30 AG066515 (PI Victor Henderson, MD, MS), P30 AG072947 (PI Suzanne Craft, PhD), P30 AG072931 (PI Henry Paulson, MD, PhD), P30 AG066546 (PIs Sudha Seshadri, MD, Gladys Maestre, MD, PhD), P30 AG086401 (PI Erik Roberson, MD, PhD), P30 AG086404 (PI Gary Rosenberg, MD), P30 AG086403 (PI Angela Jefferson, PhD), P30 AG072958 (PIs Heather Whitson, MD, Gwenn Garden, MD, PhD), P30 AG072959 (PI Jagan Pillai, MD, PhD), P30 AG092752 (Ihab Hajjar, MD, MS).

\section*{Data Availability}
The Beijing Multi-Site Air Quality dataset is publicly available from the UCI Machine Learning Repository (\url{https://doi.org/10.24432/C5RK5G})~\citep{zhang2017cautionary}.
NACC data are available upon request through \url{https://naccdata.org/data-request-process/}.

\section*{Code Availability}
The FUSHAP implementation is available at \url{https://github.com/siqili0325/FUSHAP}.

\newpage
\bibliographystyle{plainnat}
\bibliography{references}

\newpage
\appendix

\section{Summary of the FUSHAP Algorithm}
\label{app:algorithm}
\begin{algorithm}[!htbp]
\caption{FUSHAP: Variance-Reduced Shapley Attribution}
\label{alg:fushap}
\begin{algorithmic}[1]
\REQUIRE $\mathcal{LC}$, $\{\mathcal{LM}_r\}_{r=1}^R$,
  $\mathcal{UC}$, model $f$, loss $\ell$,
  significance level $\alpha$
\medskip
\STATE \textbf{Step 1: Preliminary estimator}
\STATE Estimate density ratio
  $\widehat{w}(\mathbf{x}) =
  \widehat{p}_{\mathcal{UC}}(\mathbf{x}) /
  \widehat{p}_{\mathcal{LC}}(\mathbf{x})$
  via gradient-boosted classifier on
  $\mathcal{LC} \cup \mathcal{UC}$
\STATE Sample $m$ coalitions
  $\mathbf{z}_1, \ldots, \mathbf{z}_m
  \sim \mu_{\mathrm{Sh}}$
\STATE Compute importance-weighted value functions
  $\widehat{\mathcal{V}}(\mathbf{z}_j) =
  n^{-1}\sum_{i \in \mathcal{LC}}
  \widehat{w}(\mathbf{x}_i)\,
  \eta(\mathbf{z}_j, \mathbf{x}_i, y_i)$
\STATE Solve WLS via~\eqref{eq:kkt}
  $\;\to\; \widetilde{\boldsymbol\phi}$
\medskip
\STATE \textbf{Step 2: Influence function}
\STATE Compute $\Sigma^{-1}$ from the Shapley kernel
\STATE Compute weighted residuals
  $\epsilon_w(\mathbf{z}_j, \mathbf{x}_i, y_i) =
  \widehat{w}(\mathbf{x}_i)\,
  \eta(\mathbf{z}_j, \mathbf{x}_i, y_i) -
  \widehat{\mathcal{V}}(\mathbf{z}_j)$
\STATE Compute
  $\mathbf{g}_w(\mathbf{x}_i, y_i) =
  m^{-1}\sum_{j=1}^{m}\mathbf{z}_j\,
  \epsilon_w(\mathbf{z}_j, \mathbf{x}_i, y_i)$
\STATE Compute
  $\lambda_w(\mathbf{x}_i, y_i)$
  via~\eqref{eq:lambda}
\STATE Compute
  $\widetilde{\boldsymbol\psi}_w(\mathbf{x}_i, y_i)
  = \Sigma^{-1}[\mathbf{g}_w(\mathbf{x}_i, y_i)
  + \lambda_w(\mathbf{x}_i, y_i)\,\mathbf{1}]$
\medskip
\STATE \textbf{Step 3: Control variates}
\FOR{$r = 1, \ldots, R$}
  \STATE Estimate per-site density ratio
    $\widehat{w}_r(\mathbf{x}_{\Gamma_r})$ via
    classifier on $\mathcal{LM}_r \cup \mathcal{UC}$
    using features $\Gamma_r$
  \STATE Cross-fit
    $\widehat{\boldsymbol\tau}_r$: ridge regression of
    $\widetilde{\boldsymbol\psi}$ on
    $\mathrm{poly}(\mathbf{X}_{\Gamma_r}, Y)$ with
    5-fold CV on $\mathcal{LC}$
  \STATE Refit on all $\mathcal{LC}$; evaluate
    $\widehat{\boldsymbol\tau}_r$ on $\mathcal{LM}_r$
\ENDFOR
\medskip
\STATE \textbf{Step 4: Screening}
\FOR{$r = 1, \ldots, R$}
  \STATE Compute studentized $T_r$ via~\eqref{eq:screening-stat}; assess significance by permutation test ($B_{\mathrm{perm}} = 1{,}000$)
  \IF{$\hat{p}_r < \alpha$}
    \STATE Exclude site $r$
  \ENDIF
\ENDFOR
\medskip
\STATE \textbf{Step 5: Calibration}
\FOR{$j = 1, \ldots, p$}
  \STATE Compute $A_j$ and $\mathbf{b}_j$
    via~\eqref{eq:calib-A}--\eqref{eq:calib-b}
  \STATE Add adaptive ridge:
    $A_j \gets A_j + \lambda_{\mathrm{reg}}I_R$
\ENDFOR
\STATE Solve
  $\boldsymbol\delta^* =
  \big(\sum_{j=1}^{p} A_j\big)^{-1}
  \big(\sum_{j=1}^{p} \mathbf{b}_j\big)$
  via~\eqref{eq:delta-total}
\medskip
\STATE \textbf{Output:}
  $\widehat{\boldsymbol\phi}_{\mathrm{aug}}
  = \widetilde{\boldsymbol\phi}
  + \sum_{r \in \mathrm{aligned}}
  \delta_r^*
  \big\{\bar{\boldsymbol\tau}_r^{\,\mathcal{LM}}
  - \bar{\boldsymbol\tau}_r^{\,\mathcal{LC}}\big\}$
\end{algorithmic}
\end{algorithm}

\section{Derivation of the Influence Function}
\label{app:IF-derivation}

The population Shapley vector~\eqref{eq:kkt} is a function of $(\mathbf{b}, c)$:
\begin{equation}
  \bar{\boldsymbol\phi}
  = h(\mathbf{b}, c)
  = \Sigma^{-1}
  \bigg[
    \mathbf{b}
    + \frac{c - \mathbf{1}^\top\Sigma^{-1}\mathbf{b}}
           {\mathbf{1}^\top\Sigma^{-1}\mathbf{1}}
    \,\mathbf{1}
  \bigg],
  \label{eq:app-kkt}
\end{equation}
where $\Sigma = \mathbb{E}_{\mu_{\mathrm{Sh}}} [\mathbf{z}\mathbf{z}^\top]$ depends only on~$p$ and the Shapley kernel. 
For notational convenience, define
\begin{equation}
  \mathbf{s} = \Sigma^{-1}\mathbf{1},
  \qquad
  \kappa = \mathbf{1}^\top\mathbf{s}
         = \mathbf{1}^\top\Sigma^{-1}\mathbf{1}.
  \label{eq:app-shorthand}
\end{equation}
Expanding~\eqref{eq:app-kkt}:
\begin{align}
  h(\mathbf{b}, c)
  &= \Sigma^{-1}\mathbf{b}
     + \frac{c}{\kappa}\,\mathbf{s}
     - \frac{\mathbf{s}^\top\mathbf{b}}{\kappa}\,\mathbf{s}
  \notag\\[4pt]
  &= \underbrace{
       \bigg(\Sigma^{-1}
       - \frac{\mathbf{s}\mathbf{s}^\top}{\kappa}\bigg)
     }_{M_1}\,\mathbf{b}
     \;+\;
     \underbrace{
       \frac{\mathbf{s}}{\kappa}
     }_{M_2}\,c.
  \label{eq:app-linear}
\end{align}
Since $M_1$ and $M_2$ are constant matrices (depending only on~$\Sigma$), the map $h$ is linear in~$(\mathbf{b}, c)$. 
The first-order expansion is therefore exact with no higher-order remainder.
The partial derivatives follow from~\eqref{eq:app-linear}:
\begin{align}
  \frac{\partial h}{\partial \mathbf{b}}
  &= M_1
  = \Sigma^{-1}
    - \frac{\mathbf{s}\mathbf{s}^\top}{\kappa},
  \label{eq:app-dh-db}
  \\[6pt]
  \frac{\partial h}{\partial c}
  &= M_2
  = \frac{\mathbf{s}}{\kappa}
  = \frac{\Sigma^{-1}\mathbf{1}}
         {\mathbf{1}^\top\Sigma^{-1}\mathbf{1}}.
  \label{eq:app-dh-dc}
\end{align}

Recall that $\mathbf{b}$ is defined as a weighted average over coalitions
$  \mathbf{b}
  = \mathbb{E}_{\mu_{\mathrm{Sh}}}
    \big[\mathbf{z}\big(
      \mathcal{V}(\mathbf{z}) - \mathcal{V}(\mathbf{0})
    \big)\big]$. 
When each $\mathcal{V}(\mathbf{z})$ is perturbed by $\epsilon(\mathbf{z}, \mathbf{x}, y)$ (including the null coalition $\mathcal{V}(\mathbf{0})$ which is perturbed by $\epsilon(\mathbf{0}, \mathbf{x}, y)$), the perturbed~$\mathbf{b}$ becomes
\begin{align}
  \mathbf{b} + \delta\mathbf{b}
  &= \mathbb{E}_{\mu_{\mathrm{Sh}}}
     \Big[\mathbf{z}\Big(
       \big(\mathcal{V}(\mathbf{z})
         + \epsilon(\mathbf{z}, \mathbf{x}, y)\big)
       - \big(\mathcal{V}(\mathbf{0})
         + \epsilon(\mathbf{0}, \mathbf{x}, y)\big)
     \Big)\Big]
  \notag\\[4pt]
  &= \underbrace{
       \mathbb{E}_{\mu_{\mathrm{Sh}}}
       \big[\mathbf{z}\big(
         \mathcal{V}(\mathbf{z})
         - \mathcal{V}(\mathbf{0})
       \big)\big]
     }_{\mathbf{b}}
     \;+\;
     \mathbb{E}_{\mu_{\mathrm{Sh}}}
     \big[\mathbf{z}\big(
       \epsilon(\mathbf{z}, \mathbf{x}, y)
       - \epsilon(\mathbf{0}, \mathbf{x}, y)
     \big)\big].
  \notag
\end{align}
Hence
\begin{equation}
  \delta\mathbf{b}
  = \mathbb{E}_{\mu_{\mathrm{Sh}}}
    \big[\mathbf{z}\big(
      \epsilon(\mathbf{z}, \mathbf{x}, y)
      - \epsilon(\mathbf{0}, \mathbf{x}, y)
    \big)\big].
  \label{eq:app-delta-b-raw}
\end{equation}
Since $\epsilon(\mathbf{0}, \mathbf{x}, y) = \eta(\mathbf{0}, \mathbf{x}, y) -
\mathcal{V}(\mathbf{0})$ does not depend on the coalition~$\mathbf{z}$, it is a scalar that factors out of the expectation:
\begin{align}
  \delta\mathbf{b}
  &= \underbrace{
       \mathbb{E}_{\mu_{\mathrm{Sh}}}
       \big[\mathbf{z}\,
         \epsilon(\mathbf{z}, \mathbf{x}, y)\big]
     }_{\mathbf{g}(\mathbf{x}, y)}
     \;-\;
     \underbrace{
       \mathbb{E}_{\mu_{\mathrm{Sh}}}[\mathbf{z}]
     }_{\bar{\mathbf{z}}}
     \,\epsilon(\mathbf{0}, \mathbf{x}, y).
  \label{eq:app-delta-b}
\end{align}
Similarly, from $c = \mathcal{V}(\mathbf{1}) - \mathcal{V}(\mathbf{0})$, the perturbation of~$c$ is
\begin{align}
  \delta c
  &= \big(\mathcal{V}(\mathbf{1})
       + \epsilon(\mathbf{1}, \mathbf{x}, y)\big)
     - \big(\mathcal{V}(\mathbf{0})
       + \epsilon(\mathbf{0}, \mathbf{x}, y)\big)
     - \underbrace{\big(\mathcal{V}(\mathbf{1})
       - \mathcal{V}(\mathbf{0})\big)}_{c}
  \notag\\[4pt]
  &= \epsilon(\mathbf{1}, \mathbf{x}, y)
     - \epsilon(\mathbf{0}, \mathbf{x}, y)
  \notag\\[4pt]
  &= \big[\eta(\mathbf{1}, \mathbf{x}, y)
     - \mathcal{V}(\mathbf{1})\big]
     - \big[\eta(\mathbf{0}, \mathbf{x}, y)
     - \mathcal{V}(\mathbf{0})\big]
  \notag\\[4pt]
  &= \big[\eta(\mathbf{1}, \mathbf{x}, y)
     - \eta(\mathbf{0}, \mathbf{x}, y)\big]
     - \underbrace{\big[\mathcal{V}(\mathbf{1})
     - \mathcal{V}(\mathbf{0})\big]}_{= c}.
  \label{eq:app-delta-c}
\end{align}

Define the influence function $\bar{\boldsymbol\psi}(\mathbf{x}, y)$ as the per-observation contribution to the estimation error $\widetilde{\boldsymbol\phi} -\bar{\boldsymbol\phi}$, so that
\[
  \widetilde{\boldsymbol\phi} - \bar{\boldsymbol\phi}
  = \frac{1}{n}\sum_{i \in \mathcal{LC}}
    \bar{\boldsymbol\psi}(\mathbf{x}_i, y_i).
\]
Since $h(\mathbf{b}, c) = M_1\mathbf{b} + M_2 c$ is linear, each observation's contribution is
\begin{equation}
  \bar{\boldsymbol\psi}(\mathbf{x}, y)
  = M_1\,\delta\mathbf{b}(\mathbf{x}, y)
    + M_2\,\delta c(\mathbf{x}, y)
  = \frac{\partial h}{\partial \mathbf{b}}\,
    \delta\mathbf{b}
    + \frac{\partial h}{\partial c}\,\delta c.
  \label{eq:app-chain-rule}
\end{equation}
Substituting~\eqref{eq:app-dh-db}--\eqref{eq:app-delta-c} and writing $\epsilon_0 = \epsilon(\mathbf{0}, \mathbf{x}, y)$:
\begin{align}
  \bar{\boldsymbol\psi}
  &= \bigg(\Sigma^{-1}
     - \frac{\mathbf{s}\mathbf{s}^\top}{\kappa}\bigg)
     \big(
       \mathbf{g} - \bar{\mathbf{z}}\,\epsilon_0
     \big)
     + \frac{\mathbf{s}}{\kappa}\,\delta c
  \notag\\[4pt]
  &= \Sigma^{-1}
     \big(\mathbf{g} - \bar{\mathbf{z}}\,\epsilon_0\big)
     - \frac{\mathbf{s}\,\mathbf{s}^\top
       (\mathbf{g} - \bar{\mathbf{z}}\,\epsilon_0)}
            {\kappa}
     + \frac{\mathbf{s}\,\delta c}{\kappa}
  \notag\\[4pt]
  &= \Sigma^{-1}
     \bigg[
       \mathbf{g} - \bar{\mathbf{z}}\,\epsilon_0
       + \frac{\delta c
         - \mathbf{s}^\top
           (\mathbf{g} - \bar{\mathbf{z}}\,\epsilon_0)}
              {\kappa}\,\mathbf{1}
     \bigg],
  \label{eq:app-psi-expanded}
\end{align}
where the last equality uses $\mathbf{s} = \Sigma^{-1}\mathbf{1}$ to factor $\Sigma^{-1}$ from the second and third terms.

The Shapley kernel $\mu_{\mathrm{Sh}}(\mathbf{z})$ depends on~$\mathbf{z}$ only through $|\mathbf{z}| = \mathbf{1}^\top\mathbf{z}$. 
Since each feature appears symmetrically across all coalitions of a given size,
\begin{equation}
  \bar{\mathbf{z}}
  = \mathbb{E}_{\mu_{\mathrm{Sh}}}[\mathbf{z}]
  = \tfrac{1}{2}\,\mathbf{1}.
  \label{eq:app-zbar}
\end{equation}
Substituting into~\eqref{eq:app-psi-expanded}, the expression inside the brackets becomes
\[
  \mathbf{g}
  \;-\;
 \tfrac{1}{2}\mathbf{1}\,
    \epsilon_0
  \;+\;
  \frac{\delta c
    - \mathbf{s}^\top
      (\mathbf{g} - \tfrac{1}{2}\mathbf{1}\,
       \epsilon_0)}
       {\kappa}\,\mathbf{1}.
\]
Since $-\tfrac{1}{2}\mathbf{1}\,\epsilon_0$ is proportional to~$\mathbf{1}$, it can be merged with the last term.
To do so, first expand $\mathbf{s}^\top(\mathbf{g} -
\tfrac{1}{2}\mathbf{1}\,\epsilon_0)$:
\begin{equation}
  \mathbf{s}^\top
  \big(\mathbf{g}
    - \tfrac{1}{2}\mathbf{1}\,\epsilon_0\big)
  = \mathbf{s}^\top\mathbf{g}
    - \tfrac{1}{2}
      \underbrace{(\mathbf{s}^\top\mathbf{1})}_{=\,\kappa}
      \,\epsilon_0
  = \mathbf{s}^\top\mathbf{g}
    - \tfrac{\kappa}{2}\,\epsilon_0.
  \label{eq:app-st-inner}
\end{equation}
Now collect all scalar multiples of~$\mathbf{1}$ from both terms and denote their sum by~$\lambda^*$:
\begin{align}
  \lambda^*
  &= -\tfrac{1}{2}\,\epsilon_0
     \;+\;
     \frac{\delta c
       - \mathbf{s}^\top\mathbf{g}
       + \tfrac{\kappa}{2}\,\epsilon_0}
          {\kappa}
  \notag\\[6pt]
  &= -\tfrac{1}{2}\,\epsilon_0
     \;+\;
     \frac{\delta c
       - \mathbf{s}^\top\mathbf{g}}{\kappa}
     \;+\;
     \frac{\tfrac{\kappa}{2}\,\epsilon_0}{\kappa}
  \notag\\[6pt]
  &= \frac{\delta c
       - \mathbf{s}^\top\mathbf{g}}{\kappa}.
  \label{eq:app-lambda-simplified}
\end{align}
Since all $\epsilon_0$ terms have canceled, the non-$\mathbf{1}$ part of~\eqref{eq:app-psi-expanded} reduces to~$\mathbf{g}$ alone.
Expanding $\delta c$ and $\mathbf{s}^\top\mathbf{g}$ in~\eqref{eq:app-lambda-simplified} using~\eqref{eq:app-delta-c} and $\mathbf{s}^\top = \mathbf{1}^\top\Sigma^{-1}$:
\begin{align}
  \lambda^*
  &= \frac{\delta c
       - \mathbf{s}^\top\mathbf{g}}{\kappa}
  \notag\\[4pt]
  &= \frac{
       \big[\eta(\mathbf{1}, \mathbf{x}, y)
         - \eta(\mathbf{0}, \mathbf{x}, y)\big]
       - c
       - \mathbf{1}^\top\Sigma^{-1}
         \mathbf{g}(\mathbf{x}, y)
     }{
       \mathbf{1}^\top\Sigma^{-1}\mathbf{1}
     }.
  \label{eq:app-lambda-final}
\end{align}

Substituting back into~\eqref{eq:app-psi-expanded}:
\begin{equation}
  \boxed{
  \bar{\boldsymbol\psi}(\mathbf{x}, y)
  = \Sigma^{-1}\big[
      \mathbf{g}(\mathbf{x}, y)
      + \lambda(\mathbf{x}, y)\,\mathbf{1}
    \big],}
  \label{eq:app-IF-final}
\end{equation}
with $\lambda(\mathbf{x}, y)$ given by~\eqref{eq:app-lambda-final}, recovering \eqref{eq:IF}--\eqref{eq:lambda} in the main text. 
\qed

\section{Derivation of the Calibration Weights}
\label{app:calibration}

We derive the optimal calibration weights $\boldsymbol\delta_{\cdot,j}^* = A_j^{-1}\mathbf{b}_j$ stated in~\eqref{eq:calib-A}--\eqref{eq:calib-b}.

For feature~$j$, the estimation error of the calibrated estimator~\eqref{eq:augmented-calib} is approximately
\begin{equation}
  \widehat\phi_{\mathrm{aug},j} - \bar\phi_j
  \;\approx\;
  \frac{1}{n}\sum_{i \in \mathcal{LC}}
  \bigg[\widetilde\psi_j(\mathbf{x}_i, y_i)
    - \sum_{r=1}^{R}\delta_r\,
      \widehat\tau_{r,j}(\mathbf{x}_i, y_i)\bigg]
  \;+\;
  \sum_{r=1}^{R}\frac{\delta_r}{n_r}
  \sum_{k \in \mathcal{LM}_r}
  \widehat\tau_{r,j}(\mathbf{x}_k, y_k).
  \label{eq:app-aug-error}
\end{equation}
Since $\mathcal{LC}$ and $\mathcal{LM}_r$ are independent, the variance decomposes as
\begin{equation}
  \mathrm{Var}(\widehat\phi_{\mathrm{aug},j})
  \;=\;
  \frac{1}{n}\,
  \mathrm{Var}\bigg(
    \widetilde\psi_j
    - \sum_{r}\delta_r\,\widehat\tau_{r,j}
  \bigg)
  \;+\;
  \sum_{r=1}^{R}\frac{\delta_r^2}{n_r}\,
  \mathrm{Var}_{\mathcal{LM}_r}(\widehat\tau_{r,j}).
  \label{eq:app-var-decomp}
\end{equation}
Expanding the first term:
\begin{align}
  \mathrm{Var}\bigg(
    \widetilde\psi_j
    - \sum_{r}\delta_r\,\widehat\tau_{r,j}
  \bigg)
  &= \mathrm{Var}(\widetilde\psi_j)
     - 2\sum_{r}\delta_r\,
       \mathrm{Cov}(\widetilde\psi_j,\,
         \widehat\tau_{r,j})
  \notag\\
  &\quad
     + \sum_{r}\sum_{s}\delta_r\,\delta_s\,
       \mathrm{Cov}(\widehat\tau_{r,j},\,
         \widehat\tau_{s,j}).
  \label{eq:app-var-expand}
\end{align}
Substituting~\eqref{eq:app-var-expand} into~\eqref{eq:app-var-decomp} and combining the $\delta_r^2$ terms:
\begin{align}
  V_j(\boldsymbol\delta)
  &= \frac{1}{n}\,\mathrm{Var}(\widetilde\psi_j)
     - \frac{2}{n}\sum_{r}\delta_r\,
       \mathrm{Cov}(\widetilde\psi_j,\,
         \widehat\tau_{r,j})
  \notag\\[4pt]
  &\quad
     + \frac{1}{n}\sum_{r}\sum_{s}\delta_r\,\delta_s\,
       \mathrm{Cov}(\widehat\tau_{r,j},\,
         \widehat\tau_{s,j})
     + \sum_{r}\frac{\delta_r^2}{n_r}\,
       \mathrm{Var}_{\mathcal{LM}_r}
       (\widehat\tau_{r,j}).
  \label{eq:app-var-full}
\end{align}
This is quadratic in~$\boldsymbol\delta$. Taking the derivative with respect to~$\delta_r$ and setting it to zero:
\begin{equation}
  \frac{\partial V_j}{\partial\delta_r}
  = -\frac{2}{n}\,
    \mathrm{Cov}(\widetilde\psi_j,\,
      \widehat\tau_{r,j})
    + \frac{2}{n}\sum_{s}\delta_s\,
      \mathrm{Cov}(\widehat\tau_{r,j},\,
        \widehat\tau_{s,j})
    + \frac{2\delta_r}{n_r}\,
      \mathrm{Var}_{\mathcal{LM}_r}
      (\widehat\tau_{r,j})
  = 0.
  \label{eq:app-deriv}
\end{equation}
Dividing by $2/n$ and rearranging:
\begin{equation}
  \sum_{s=1}^{R}\delta_s
  \bigg[
    \mathrm{Cov}(\widehat\tau_{r,j},\,
      \widehat\tau_{s,j})
    + \mathbf{1}_{r=s}\,\frac{n}{n_r}\,
      \mathrm{Var}_{\mathcal{LM}_r}
      (\widehat\tau_{r,j})
  \bigg]
  = \mathrm{Cov}(\widetilde\psi_j,\,
      \widehat\tau_{r,j}).
  \label{eq:app-normal-eq}
\end{equation}
This is the linear system $A_j\boldsymbol\delta_{\cdot,j} = \mathbf{b}_j$ with
\begin{align}
  [A_j]_{rs}
  &= \mathrm{Cov}_{\mathcal{LC}}
     (\widehat\tau_{r,j},\,\widehat\tau_{s,j})
     + \mathbf{1}_{r=s}\,\frac{n}{n_r}\,
       \mathrm{Var}_{\mathcal{LM}_r}
       (\widehat\tau_{r,j}),
  \notag\\[4pt]
  [\mathbf{b}_j]_r
  &= \mathrm{Cov}_{\mathcal{LC}}
     (\widetilde\psi_j,\,\widehat\tau_{r,j}),
  \notag
\end{align}
recovering~\eqref{eq:calib-A}--\eqref{eq:calib-b} in the main text. 
When a single per-site scalar $\delta_r$ is used across all features, the total variance $\sum_{j=1}^{p} V_j(\boldsymbol\delta)$ is minimized by summing~\eqref{eq:app-normal-eq} over $j$, yielding $\big(\sum_j A_j\big)\boldsymbol\delta = \sum_j \mathbf{b}_j$.
\qed

\section{Additional Experimental Results}
\label{app:additional}

\subsection{Simulation Experiment~3: Detailed Screening Results}

Table~\ref{tab:exp3-detail} reports per-site rejection rates and MSE across all misalignment strengths tested in Experiment~3. 
Panel~(a) confirms that the three aligned sites maintain rejection rates between $2\%$ and $7\%$ across all values of $\Delta$, consistent with the nominal $\alpha = 0.05$. 
Panel~(b) shows that FUSHAP with screening stabilizes MSE between $0.061$ and $0.066$ across all misalignment strengths, while FUSHAP without screening degrades beyond the single-site baseline for $\Delta \geq 1$.

\begin{table}[!htbp]
\centering
\caption{Experiment~3: per-site rejection rates and MSE across misalignment strengths ($\alpha = 0.05$, $B = 100$). Aligned sites satisfy Assumption~\ref{asm:mar} exactly.}
\label{tab:exp3-detail}
\small

\medskip
\textit{(a) Rejection rates (\%)}
\medskip

\begin{tabular}{r cccc}
\toprule
$\Delta$ & Site 1 & Site 2 & Site 3 & Site 4 \\
& (aligned) & (aligned) & (aligned) & (misaligned) \\
\midrule
0    & 4  & 7  & 2  & 5 \\
0.25 & 4  & 7  & 2  & 82 \\
0.5  & 4  & 7  & 2  & 100 \\
1.0  & 4  & 7  & 2  & 100 \\
1.5  & 4  & 7  & 2  & 100 \\
2.0  & 4  & 7  & 2  & 100 \\
3.0  & 4  & 7  & 2  & 100 \\
\bottomrule
\end{tabular}

\bigskip
\textit{(b) MSE comparison}
\medskip

\begin{tabular}{r ccc}
\toprule
$\Delta$ & LC-only & FUSHAP & FUSHAP \\
& & (no screen) & (screen) \\
\midrule
0    & 0.164 & 0.061 & 0.061 \\
0.25 & 0.164 & 0.064 & 0.066 \\
0.5  & 0.164 & 0.075 & 0.066 \\
1.0  & 0.164 & 0.124 & 0.066 \\
1.5  & 0.164 & 0.187 & 0.066 \\
2.0  & 0.164 & 0.232 & 0.066 \\
3.0  & 0.164 & 0.249 & 0.066 \\
\bottomrule
\end{tabular}
\end{table}

\subsection{Beijing Air Quality}
\label{app:beijing}

The Beijing Multi-Site Air Quality dataset~\citep{zhang2017cautionary} contains hourly measurements from 12 monitoring stations over March 2013 to February 2017. We aggregate to daily averages and remove days with any missing values within each station.

The target variable is daily mean PM2.5 concentration.
The $p = 10$ predictor features are: PM10, SO2, NO2, CO, O3 (pollutants) and TEMP, PRES, DEWP, RAIN, WSPM (meteorological variables).

To construct a controlled multi-site scenario with complementary blockwise missingness, we designate one urban station (Dongsi) as $\mathcal{LC}$ with all features observed ($n = 300$ days subsampled per replication). 
Three suburban stations serve as $\mathcal{LM}$ sites, each with a different pair of features artificially removed: CO and O3 at Changping, SO2 and NO2 at Huairou, and DEWP and PRES at Shunyi.
The remaining eight stations form $\mathcal{UC}$ ($N = 11{,}375$ days) with complete feature coverage but no outcome variable used during Shapley estimation. 
This design ensures complementary missingness patterns across sites while preserving real inter-station distributional heterogeneity.

All features and the outcome are standardized using $\mathcal{UC}$ means and standard deviations. Four regression models (linear, random forest, GBM, MLP) are trained on $15{,}646$ observations from all non-$\mathcal{LC}$ stations and held fixed during Shapley estimation. 
The reference attribution vector is computed from $2{,}000$ held-out training observations.

\subsection{NACC Alzheimer's Data}
\label{app:nacc}

The National Alzheimer's Coordinating Center (NACC) Uniform Data Set (UDS) aggregates clinical assessments from over 40 Alzheimer's Disease Research Centers across the United States. 
We use the investigator dataset which contains demographic, clinical, and cognitive variables collected under heterogeneous protocols across centers.

The outcome variable is the Mini-Mental State Examination (MMSE) score, a continuous measure of cognitive function ranging from 0 (severe impairment) to 30 (no impairment). 
The $p = 9$ predictor features are: sex, race, age at visit (NACCAGE), weight, height, living situation (NACCLIVS), alcohol use history, tobacco use history (TOBAC100), and CDR language domain score (CDRLANG). 
Observations with missing values in any feature or the outcome are excluded, yielding $17{,}200$ complete cases from $34$ centers.

To construct a controlled multi-site scenario with complementary blockwise missingness, we randomly sample $n = 300$ patients from the pooled complete cases to serve as $\mathcal{LC}$. 
The remaining patients are partitioned into three $\mathcal{LM}$ sites and one $\mathcal{UC}$ pool. 
Each $\mathcal{LM}$ site has a different triplet of features artificially removed (Table~\ref{tab:nacc}), ensuring complementary missingness patterns. 
This design satisfies Assumption~\ref{asm:mar} by construction, as all partitions are drawn from the same population.

Continuous features (NACCAGE, WEIGHT, HEIGHT, CDRLANG) are standardized using the $\mathcal{UC}$ means and standard deviations.
Four regression models (ridge, random forest, GBM, MLP) are trained on all non-$\mathcal{LC}$ patients ($n_{\mathrm{train}} = 16{,}900$) and held fixed during Shapley estimation.
The reference attribution vector is computed from all non-$\mathcal{LC}$ patients ($n_{\mathrm{train}} = 16{,}900$) with $m = 300$ sampled coalitions, providing an evaluation target independent of the $\mathcal{LC}$ sample used for attribution estimation.

\end{document}